\documentclass{article}

\usepackage[preprint]{neurips_2025}

\usepackage[utf8]{inputenc}
\usepackage[T1]{fontenc}
\usepackage{amsmath}
\usepackage{url}
\usepackage{xurl}
\usepackage[hidelinks]{hyperref}
\usepackage[capitalize,noabbrev]{cleveref}
\usepackage{booktabs}
\usepackage{multirow}
\usepackage{array}
\usepackage{threeparttable}

\usepackage{graphicx}
\usepackage{caption}
\usepackage{subcaption}
\usepackage{wrapfig}

\usepackage{amsfonts}

\usepackage{amssymb}
\usepackage{mathtools}
\usepackage{amsthm}

\usepackage{algorithm}
\usepackage{algorithmic}

\usepackage{nicefrac}
\usepackage{microtype}
\usepackage{pifont}
\usepackage{enumitem}

\usepackage[table]{xcolor}

\newcommand{\oursacc}[2]{%
  \begin{tabular}[t]{@{}c@{}}
    #1\\[-3pt]
    {\fontsize{7}{7}\selectfont #2}
  \end{tabular}%
}

\theoremstyle{plain}

\theoremstyle{definition}

\theoremstyle{remark}

\title{Fusion Anything: A Generalized Multimodal Foundation Model}

\author{
\normalfont
\large
Huizi Cui$^{1,\dagger}$,
Zongbo Han$^{2,\dagger,\ddagger}$,
Chenggong Ding$^{1}$,
Naichuan Xiao$^{1}$,
Jialong Yang$^{1}$
\\[1pt]
\large
Jingdong Chen$^{1}$,
Yafei Yang$^{1}$,
Guangyu Wang$^{2}$,
Qinghua Hu$^{1}$,
Changqing Zhang$^{1,*}$
\\[3pt]
{\normalsize
$^{1}$Tianjin University
\qquad
$^{2}$Beijing University of Posts and Telecommunications
}
}

\begin{document}

\maketitle

\begingroup
\renewcommand{\thefootnote}{}
\footnotetext{%
\fontsize{8pt}{9.5pt}\selectfont
\makebox[\linewidth][l]{%
\textsuperscript{$\dagger$}\,Equal contribution.
\textsuperscript{$\ddagger$}\,Project lead: hanzongbo.mail@gmail.com
\textsuperscript{$*$}\,Corresponding author: zhangchangqing@tju.edu.cn
}}
\endgroup

\begin{abstract}
Making prediction with multimodal data is widely used in diverse scenarios. Existing multimodal fusion models, once deployed, can only handle predefined modalities (e.g., vision, text and audio) and single task, making it difficult to quickly adapt to new downstream applications. Therefore, a natural yet aggressive question arises - whether there exists a general multimodal fusion model that can be applied to arbitrary modality combinations and arbitrary prediction tasks. We argue that a unified multimodal fusion model should not depend on specific modalities and should instead encode transferable patterns of multimodal correlation. To this end, we propose a simple and effective learning paradigm based on training on large-scale synthetic multimodal datasets generated with Structural Multimodal Causal Models (SMCMs), which formally characterizes the generative processes of real-world multimodal data. Building on this framework, we propose the \textbf{Fusion Anything Model (FAM)}, a foundation model for generalized multimodal data fusion. By constructing large-scale synthetic multimodal data with diverse correlation patterns, our model encodes transferable multimodal correlations during training and activates appropriate associations through in-context examples during inference. Extensive experiments on $18$ real-world datasets spanning $12$ modalities and $11$ prediction tasks demonstrate that our model achieves competitive performance with specialized models without task-specific adaptation.
\end{abstract}

\section{Introduction}
Making prediction with multimodal data focuses on exploiting synergy across modalities, enabling models to capture richer information and make better predictions \cite{baltruvsaitis2018multimodal, xu2023multimodal, zhang2026multimodal}. By modeling interactions among heterogeneous modalities, multimodal learning has become an important paradigm for understanding complex real-world data \cite{, liang2023quantifying}. However, existing multimodal fusion models are predominantly developed under a task-specific paradigm, where they are designed for predefined modalities and one specific prediction objective \cite{haque2020illuminating, zhang2023provable, wei2020multi}. Consequently, these models mainly learn fusion strategies tailored to specific scenarios, making it difficult to capture generalized multimodal correlation patterns beyond predefined modality combinations and task settings.

\begin{figure}[htbp]
\centering
\includegraphics[width=0.97\columnwidth]{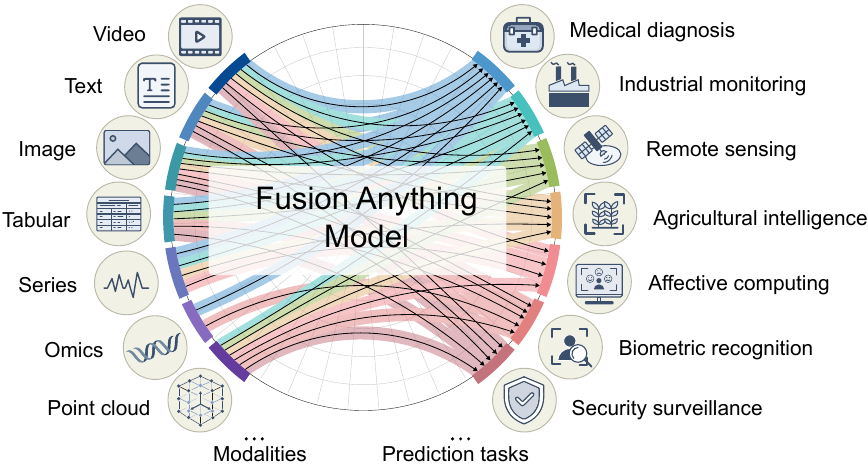}
\caption{Fusion Anything Model for generalized multimodal prediction. A unified foundation model accommodates arbitrary modality combinations and diverse prediction tasks, enabling direct prediction without task-specific parameter adaptation.}
\label{fusion}
\end{figure}

Recent success of foundation models demonstrates that large-scale pretraining over diverse data distributions enables models to learn transferable patterns and acquire generalizable capabilities across diverse tasks and domains \cite{bommasani2021opportunities,  kirillov2023segment, yang2024depth, hollmann2025accurate, wang2026limix}. Following this paradigm, multimodal foundation models have emerged to integrate heterogeneous information sources and advance multimodal understanding and prediction \cite{liang2024foundations, wang2026multimodal, dong2026advances}. However, existing models remain largely confined to a limited modality space, focusing on several well-established modalities such as vision, language, and audio while leaving many other heterogeneous modalities largely unexplored. Such modality-specific designs hinder their ability to capture the complex correlations of generalized modalities that naturally arise in real-world scenarios. Accordingly, a fundamental and aggressive question arises: \textbf{can we develop a multimodal foundation model that can rapidly adapt to arbitrary modality combinations and prediction tasks (see Figure~\ref{fusion})}?

The first challenge of achieving such a generalized multimodal foundation model is the requirement of a large-scale and diverse collection of multimodal data that sufficiently cover the characteristics of real-world data. However, constructing such a collection of real data is inherently challenging. Existing multimodal datasets are highly concentrated on limited modality combinations \cite{xue2026multibench++, liang2021multibench}, while many other modalities remain scarce due to domain-specific acquisition processes and limited data permission. Moreover, collecting, annotating, and aligning large-scale datasets across diverse domains introduces substantial cost and practical difficulties. These limitations make it difficult to obtain comprehensive training data that cover diverse modality combinations and multimodal correlation patterns. Another challenge lies in the model design, namely how to develop a unified architecture that can adapt to future unseen prediction tasks. Existing multimodal models are predominantly optimized for predefined modalities and task objectives \cite{wang2026multimodal, han2022multimodal}, resulting in architectures that are tightly coupled with specific scenarios. Such designs limit their ability to transfer learned multimodal correlation patterns to new tasks and modality combinations.

To address these challenges, in this paper, we propose a new learning paradigm for generalized multimodal fusion based on large-scale synthetic multimodal data. Instead of relying on real multimodal data with limited modality coverage, we construct synthetic multimodal data with Structural Multimodal Causal Models (SMCMs), where complex multimodal correlations naturally emerge through the underlying data generation processes. By training over a broad collection of synthetic multimodal data, our model learns transferable multimodal correlation patterns beyond specific modality combinations and datasets. Building upon this paradigm, we develop Fusion Anything Model (FAM), a multimodal foundation model with the in-context learning capability~\cite{hollmann2025accurate} for diverse multimodal prediction tasks. Through a unified token generation mechanism, our model maps heterogeneous modalities into a unified token space and enables multimodal in-context learning over diverse correlation patterns. During inference, it leverages contextual examples from real datasets to activate task-relevant multimodal correlations and perform prediction without additional parameter updates. Extensive experiments across diverse real-world datasets demonstrate that our Fusion Anything Model outperforms the specialized multimodal fusion models while substantially improving generalization beyond predefined scenarios. Our contributions are summarized as follows:

\begin{itemize}
    \item We introduce a new learning paradigm for generalized multimodal fusion by constructing large-scale synthetic datasets with diverse SMCMs, enabling general multimodal correlation patterns are pretrained in the fundation model.
    
    \item We propose the Fusion Anything Model, a unified foundation model that adapts to diverse modality combinations and prediction tasks through multimodal in-context learning.
    
    \item Extensive experiments on diverse real-world datasets demonstrate that the proposed model achieves promising performance compared with specialized models without task-specific adaptation.
\end{itemize}

\section{Related Work}

\textbf{Multimodal Fusion.}
Multimodal fusion aims to integrate heterogeneous information from multiple modalities for more comprehensive understanding \cite{zhang2019cpm}. Early research mainly explores feature-level and decision-level fusion strategies. Feature-level fusion combines modality representations before prediction \cite{ramachandram2017deep, atrey2010multimodal}, while decision-level fusion aggregates predictions from individual modalities, providing modularity but limited cross-modal interaction \cite{han2022trusted, zhang2023provable}. With the development of deep learning, intermediate fusion methods introduce separate encoders for each modality and interaction modules to jointly learn modality representations and cross-modal correlations \cite{xu2023multimodal}. Recently, Transformer-based architectures have become the dominant paradigm due to their ability to model complex dependencies through attention mechanisms \cite{liang2021multibench, girdhar2022omnivore, wang2022multimodal}. Existing approaches typically follow either single-stream architectures, which project multiple modalities into a shared representation space, or multi-stream architectures, which retain separate encoders for each modality and perform cross-attention fusion \cite{nagrani2021attention}. Furthermore, cross-attention mechanisms have been widely adopted to explicitly capture interactions between heterogeneous modalities and dynamically integrate complementary information \cite{sun2021multimodal, zhan2021product1m}. Despite their effectiveness on specific tasks, these methods usually assume predefined modality combinations and require task-specific architectures, limiting their generalization to diverse and unseen multimodal scenarios.

\textbf{Multimodal Foundation Models.}
Recent advances in foundation models have shifted multimodal learning toward large-scale pretraining and transferable representations \cite{huang2026dissecting, dong2026advances}. Early multimodal foundation models mainly focus on learning aligned representations between vision and language, with representative works such as CLIP \cite{radford2021learning} and ALIGN \cite{jia2021scaling}. Building upon these advances, multimodal large language models (MLLMs) further integrate visual encoders with large language models to enable multimodal understanding and reasoning, including Flamingo \cite{alayrac2022flamingo}, BLIP-2 \cite{li2023blip}, and LLaVA \cite{liu2023visual}. More recent studies have explored unified multimodal architectures by mapping different modalities into shared token spaces and leveraging autoregressive modeling, such as Chameleon \cite{team2024chameleon}, Emu3 \cite{wang2026multimodal}, and Janus \cite{wu2025janus}. However, existing works are still primarily designed for limited modality combinations and often rely on extensive modality-specific pretraining, making it challenging to achieve generalization across arbitrary modalities and task settings.

\textbf{Tabular Foundation Models.}
TabPFN \cite{hollmann2025accurate} is a tabular foundation model pre-trained on large-scale synthetic datasets with diverse causal structures, enabling it to capture general variable dependencies and perform prediction on unseen structured datasets through contextual examples. Subsequent work has further advanced this line of research \cite{lee2026mitigating, ma2026tabdpt}. For example, LimiX \cite{wang2026limix} extends structured-data modeling to multiple tasks within a unified model, including classification, regression, imputation, and data generation. MMPFN \cite{kim2026multimodalpfn} further incorporates image and text representations into TabPFN through modality projector, but requires fine-tuning on individual real-world datasets. Overall, current tabular foundation models are largely centered on structured-data prediction, while extending such transferable modeling capabilities to heterogeneous multimodal data remains underexplored.

\section{Method}

\subsection{Preliminaries}
Let $\mathcal{D}=\{(\mathbf{X}_i,y_i)\}_{i=1}^{N}$ be a dataset of $N$ samples with $M$ modalities, where $\mathbf{X}_i=(\mathbf{x}_i^1,\ldots,\mathbf{x}_i^M)$, $\mathbf{x}_i^m\in\mathbb{R}^{d_m}$ is the encoded representation of modality $m$, and $y_i$ is the target label. Existing fusion methods typically fit a model $p_{\theta^*}$ to each dataset, learning correlations specific to its modality combination and prediction task. In this work, we aim to learn a generalized foundation model $p_\theta$ which captures transferable multimodal correlation patterns and is applicable to novel multimodal prediction tasks.

\subsection{Generation of Synthetic Multimodal Data}
\begin{figure}[htbp]
\centering
\includegraphics[width=0.99\columnwidth]{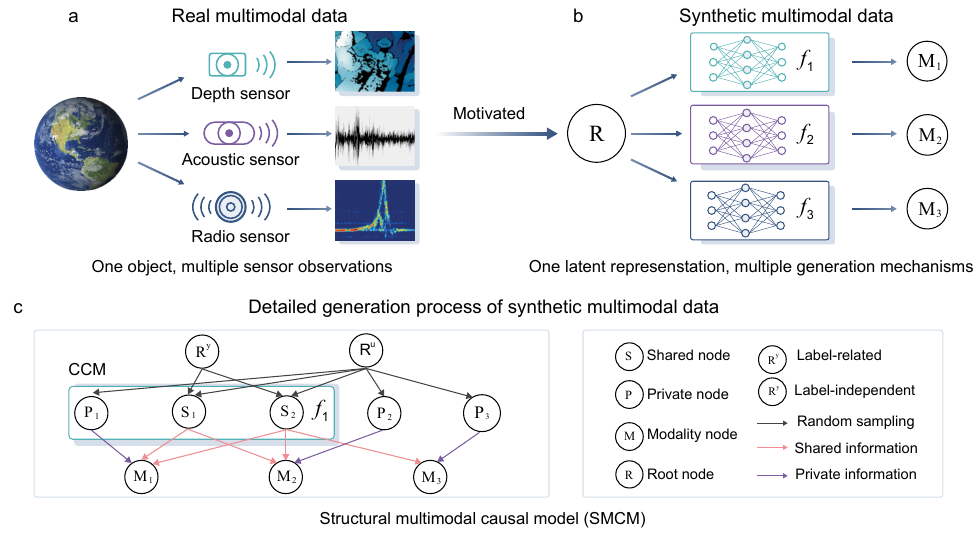}
\caption{Synthetic multimodal dataset generation. (a) Real-world multimodal observations are produced by different sensors perceiving the same underlying object. (b) Motivated by this process, we generate synthetic modalities from a common latent representation through distinct generation mechanisms. (c) An example SMCM with three modalities $M_1$, $M_2$, and $M_3$. The root representation comprises label-related and label-independent components, and the CCM controls the allocation of latent information through shared and private nodes. Shared nodes provide common information to multiple modalities, while each private node supplies modality-specific variation to its corresponding modality. Varying these node configurations and connections induces diverse multimodal correlation patterns.}
\label{data}
\end{figure}

Real-world multimodal datasets are limited and costly to collect. Therefore, we construct synthetic datasets with controllable modality combinations and correlation patterns to support the model training across diverse multimodal settings. Figure~\ref{data} illustrates the motivation and overview of our synthetic multimodal data generation process. As shown in Figure~\ref{data}(a), real-world data are commonly collected by observations of a single target by different sensors. Inspired by this observation, Figure~\ref{data}(b) depicts the basic principle adopted in our synthetic construction: a common root node $R$ is connected to multiple modality nodes $M_m$ through distinct generation mechanisms $f_m$. At the sample level, these mechanisms transform the underlying latent information into heterogeneous modality representations, naturally inducing cross-modal correlations through their common source. 

However, directly generating each modality from the root provides little control over how information is shared across modalities or retained within individual modalities, making it difficult to cover diverse multimodal correlation patterns with limited generated data. To address this limitation, we propose Structural Multimodal Causal Models (SMCMs) illustrated in Figure~\ref{data}(c). An SMCM characterizes the multimodal generation process through a structured causal graph comprising a root node, a Correlation Control Module (CCM), and multiple modality nodes. The CCM regulates how latent information contributes to different modalities through shared nodes $S_j$ and private nodes $P_m$, shared nodes convey selected information to multiple modality nodes, while each private node contributes modality-specific variation only to its associated modality. Varying the configurations of these nodes and their connections to modality nodes enables the generation of synthetic datasets with diverse correlation patterns.

\textbf{Generation Mechanisms in SMCM.}
The SMCM organizes multimodal data generation through four types of nodes: a root node $R$, shared nodes $S_j$, private nodes $P_m$, and modality nodes $M_m$. To define the internal generation mechanisms of the root and modality nodes, we use Structural Causal Models (SCMs) \cite{pearl2009causality}. An SCM consists of a directed acyclic graph and a set of structural mechanisms that specify how its variables are generated \cite{hollmann2025accurate,wang2026limix}. Within each SCM, each root variable is sampled from either a Gaussian distribution $\mathcal{N}(0,\tau_{\mathrm{init}}^2)$ or a uniform distribution $\mathcal{U}(-c_{\mathrm{init}},c_{\mathrm{init}})$, with the distribution type randomly selected and the scale parameters treated as hyperparameters. Each non-root variable $v_a$ is then generated from its parent variables as follows:
\begin{equation}
v_a =
f_a\left(
\left\{
g_{a,b}(v_b)
\right\}_{v_b \in \mathrm{Pa}(v_a)},
\epsilon_a
\right),
\end{equation}
where $\mathrm{Pa}(v_a)$ denotes the set of parent variables of $v_a$, $g_{a,b}(\cdot)$ denotes the edge function that transforms parent variable $v_b$ before its contribution to $v_a$, $f_a(\cdot)$ denotes an aggregation function that combines the transformed parent variables, and $\epsilon_a$ represents an observational noise term. The diversity of variable dependencies is further controlled by the functional forms of the edge and aggregation functions. The edge functions can be instantiated using neural networks with different activation functions (e.g., sigmoid, ReLU, and Tanh), convolutional mappings, or decision-tree functions, while the aggregation function can take the form of mean, weighted, or neural aggregation.

\textbf{Root Node.}
The root node $R$ represents an underlying (complete) representation inducing all modalities \cite{zhang2018generalized}, from which the shared, private, and modality nodes are subsequently constructed. We decompose the root representation into label-related and label-independent components to explicitly control task-related information while preserving other latent variations \cite{paige2017learning}. Specifically, for sample $i$, the root representation is defined as:
\begin{equation}
\mathbf{r}_i
=
(\mathbf{h}_i+\alpha\boldsymbol{\mu}_{y_i})\mathbf{A}
=
[\mathbf{r}^{y}_i\Vert\mathbf{r}^{u}_i],
\end{equation}
where $\mathbf{r}_i$ denotes the realization of root node $R$, $\mathbf{h}_i$ is the initial latent representation of sample $i$, $\boldsymbol{\mu}_{y_i}$ denotes the label-specific shift associated with class $y_i$, and $\alpha$ controls the strength of the label-related variation. The transformation matrix $\mathbf{A}$ aligns the label-related variation with a dedicated subspace, yielding the label-related component $\mathbf{r}^{y}_i$, while the remaining dimensions form the label-independent component $\mathbf{r}^{u}_i$. These two components are concatenated along the representation dimension, denoted by $\Vert$, to form the complete root representation $\mathbf{r}_i$.

\textbf{Shared and Private Nodes.}
The Correlation Control Module (CCM) consists of shared and private nodes and enables explicit control over the multimodal correlation structures formed during modality generation. As illustrated in Figure~\ref{data}(c), given the root representation $\mathbf{r}_i=[\mathbf{r}_i^{y}\Vert\mathbf{r}_i^{u}]$, each shared node $S_j$ selects dimensions indexed by $\Omega_j^{y}$ and $\Omega_j^{u}$ from the label-related and label-independent components, respectively. The selected dimensions are then processed through a nonlinear transformation to produce the shared representation $\mathbf{s}_i^{j}$. Each shared node is connected to multiple modality nodes. In contrast, private node $P_m$ selects dimensions indexed by $\Omega_m^{p}$ from the label-independent component $\mathbf{r}_i^{u}$ that are not assigned to shared nodes. The selected dimensions are likewise processed by an independent nonlinear transformation to produce the private representation $\mathbf{p}_i^{m}$ associated exclusively with modality $M_m$.

\textbf{Modality Node.}
Each modality node $M_m$ characterizes the specific modality representation in each synthetic dataset, which integrates information derived from the root node, subsequently the associated shared nodes and private node. For sample $i$, the representations of all shared nodes connected to $M_m$ are concatenated and denoted by $\mathbf{s}_i^{(m)}$. The representation associated with $M_m$ is defined as:
\begin{equation}
\mathbf{x}_i^m
=
\mathcal{F}_m
\left(
\mathbf{r}_i[\Omega_m^r]
\Vert
\mathbf{s}_i^{(m)}[\Omega_m^s]
\Vert
\mathbf{p}_i^m[\Omega_m^{p\prime}],
\boldsymbol{\epsilon}_i^m
\right),
\end{equation}
where $\mathbf{x}_i^m$ denotes the realization of modality node $M_m$, and $\mathcal{F}_m(\cdot)$ denotes the SCM-based structural mechanism governing the transformation from the selected root, shared, and private representations to $\mathbf{x}_i^m$. The index sets $\Omega_m^r$, $\Omega_m^s$, and $\Omega_m^{p\prime}$ specify the selected dimensions from $\mathbf{r}_i$, $\mathbf{s}_i^{(m)}$, and $\mathbf{p}_i^m$, respectively, while $\boldsymbol{\epsilon}_i^m$ denotes the noise term. By varying the graph configurations and structural mechanisms across SMCMs, we construct synthetic multimodal datasets with diverse correlation structures.

\subsection{Fusion Anything Model}
\begin{figure}[htbp]
\centering
\includegraphics[width=0.99\columnwidth]{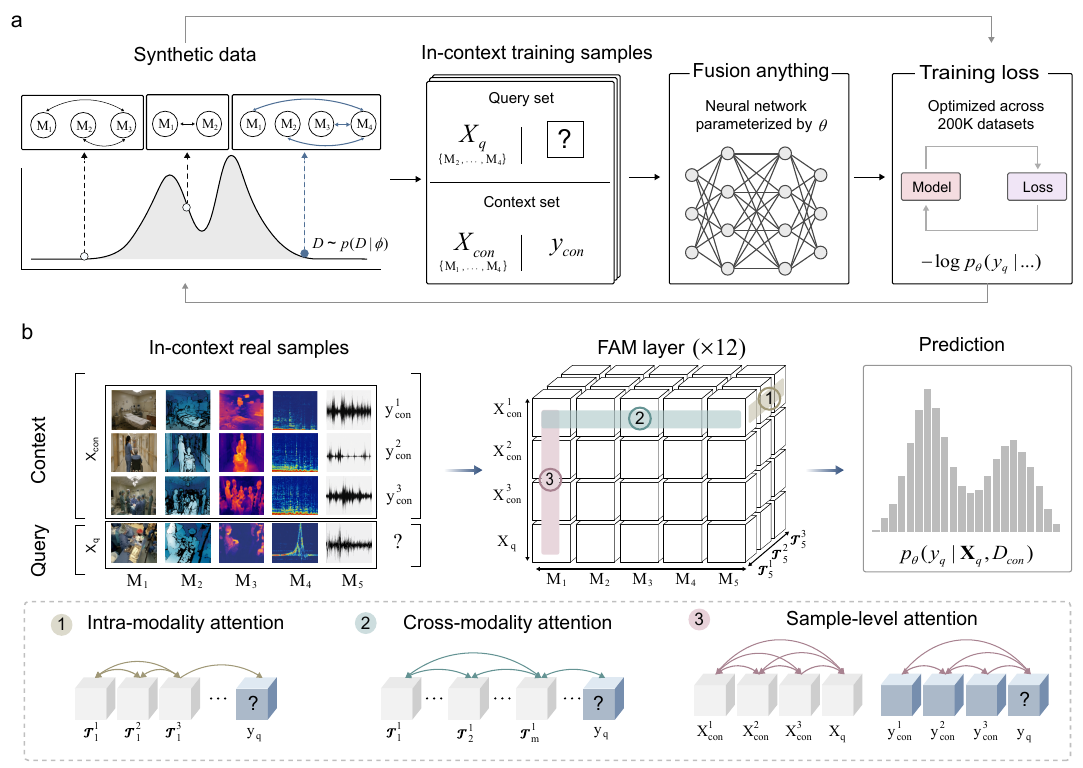}
\caption{Overview of FAM. (a) Training: synthetic datasets generated by SMCMs are organized to in-context prediction tasks, with query-label guiding the optimization of the proposed FAM across tasks. (b) Inference: given labeled context examples from a real-world multimodal task, FAM predicts labels for unlabeled query samples by leveraging intra-modality, cross-modality, and sample-level attentions.}
\label{fam}
\end{figure}

FAM is designed to learn transferable correlation patterns through in-context learning over large-scale synthetic multimodal data. Figure~\ref{fam} presents an overview of the training and inference of our model. During training, each dataset generated with SMCM is partitioned into a labeled context set $\mathcal{D}_{\mathrm{con}}$ and a query set $\mathcal{D}_{\mathrm{q}}$, and FAM is optimized to predict query labels conditioned on the context examples. Training across diverse datasets encourages FAM to encode  correlation patterns that generalize beyond individual modality combinations and prediction objectives. During inference, FAM employs labeled context examples from real-world tasks to activate the appropriate multimodal correlation patterns learned during training and predict query labels.

\textbf{Modality Projector.}
The modality projector maps heterogeneous modality representations into a common $d$-dimensional token space for joint multimodal modeling. For sample $i$, the representation $\mathbf{x}_i^m$ associated with modality $M_m$ is projected into a token sequence:
\begin{equation}
\mathcal{T}_i^m
=
\mathcal{P}(\mathbf{x}_i^m),
\end{equation}
where $\mathcal{P}(\cdot)$ denotes the modality projector and $\mathcal{T}_i^m=(\mathbf{t}_{i,\ell}^m)_{\ell=1}^{L_m}$ is a sequence of $L_m$ tokens with $\mathbf{t}_{i,\ell}^m\in\mathbb{R}^{d}$. Specifically, parallel nonlinear projection heads incorporate gated linear units (GLUs) to selectively regulate feature activations and generate candidate token representations from $\mathbf{x}_i^m$. A cross-attention module uses $L_m$ learnable query embeddings to aggregate the candidate representations into a sequence of $L_m$ modality tokens. The resulting token sequences are concatenated along the sequence dimension across all available modalities, yielding a unified multimodal representation whose length is determined by the available modalities and their respective token counts.

\textbf{Attention Mechanism.}
FAM models intra-modality, cross-modality, and sample-level correlations through attention mechanisms over the projected multimodal tokens. Intra-modality attention operates among tokens belonging to the same modality within each sample, capturing relationships among different components of its representation. Cross-modality attention enables interactions among tokens from different modalities within the same sample, integrating consistent and complementary information for prediction. Sample-level attention operates across samples, allowing query tokens to incorporate information from labeled context examples and supporting the inference of predictive relationships relevant to the current task. Together, these attention operations integrate multimodal information within samples and contextual information across samples, enabling context-conditioned multimodal prediction.

\textbf{Multimodal In-context Learning Objective.}
The proposed FAM is trained across diverse synthetic multimodal datasets with different causal structures. For each task, the labeled context set $\mathcal{D}_{\mathrm{con}}$ provides task-specific information for identifying relevant multimodal correlations, while the query set $\mathcal{D}_{\mathrm{q}}$ provides the prediction targets used for optimization. The training objective is defined as follows:
\begin{equation}
\mathcal{L}(\theta)
=
-\mathbb{E}_{\mathcal{D}\sim p(\mathcal{D})}
\left[
\frac{1}{|\mathcal{D}_{\mathrm{q}}|}
\sum_{(\mathbf{X}_i,y_i)\in\mathcal{D}_{\mathrm{q}}}
\log p_{\theta}(y_i\mid\mathbf{X}_i,\mathcal{D}_{\mathrm{con}})
\right],
\end{equation}
where $\theta$ denotes the trainable model parameters. By optimizing this objective across synthetic multimodal tasks, FAM learns to exploit contextual examples for prediction under diverse causal structures. We build on the pretrained TabPFN backbone \cite{hollmann2025accurate} to leverage its existing capacity for modeling feature dependencies and performing context-conditioned prediction. During training, the backbone is updated through Low-Rank Adaptation (LoRA) \cite{hu2021lora}, together with the modality projector and prediction head, while the remaining pretrained parameters are frozen.

\subsection{Inference}
After training on synthetic multimodal data, the proposed FAM can be directly applied to real-world datasets. As shown in Figure~\ref{fam}(b), for a real-world dataset $\mathcal{D}$ with $M$ modalities, observations from each modality are first encoded by pre-trained modality encoders to obtain $\mathbf{x}_i^m$ and subsequently mapped into the unified token space. Since our model concentrates on the modality representations rather than encoder-specific architectures, it remains flexible with respect to the choice of pre-trained encoders. Given a labeled real-world context set $\mathcal{D}_{\mathrm{con}}$ and an unlabeled query input $\mathbf{X}_q$, the model predicts its target according to $p_{\theta}(y_q\mid\mathbf{X}_q,\mathcal{D}_{\mathrm{con}})$ while keeping all model parameters fixed. This enables direct prediction on novel multimodal datasets with diverse modality combinations and prediction tasks without task-specific optimization.

\section{Experiments}

\subsection{Experimental Settings}

\textbf{Datasets and Evaluation Metrics.}
We evaluate FAM on a diverse collection of real-world multimodal datasets covering various application domains, such as social media analysis \cite{suryawanshi2020multimodal, xue2026multibench++, niu2016sentiment}, affective computing \cite{castro2019towards, xu2022met, livingstone2018ryerson, sharma2020semeval, schmidt2018introducing}, visual understanding \cite{silberman2012indoor, leiva2020enrico, song2015sun}, medical \cite{wu2023gamma, weinstein2013cancer, rotemberg2021patient, pacheco2020pad, bennett2018religious} and remote sensing \cite{xue2026multibench++, irvin2020forestnet}. The main paper presents evaluation results on $18$ datasets, while the Appendix reports the complete results across all $23$ datasets, together with detailed descriptions of their modality configurations and data splits. Performance is assessed using Accuracy (ACC) and the Area Under the Receiver Operating Characteristic Curve (AUC).

\textbf{Comparison Methods.}
We compare FAM with seven representative supervised multimodal fusion methods, categorized into early, intermediate, and late fusion according to where multimodal information is integrated. Early fusion combines modality features before joint predictive modeling, including Concat \cite{xue2026multibench++} and Early Fusion Transformer (EFT) \cite{liang2021multibench}. Intermediate fusion combines representations from different modalities within the encoding process, including Multi-to-One (MTO) \cite{li2021ai}, One-to-Multi (OTM) \cite{lin2020interbert}, and Cross-Attention Fusion (CAF) \cite{lu2019vilbert}. Late fusion combines outputs from individual modality predictors, represented by Logit Summation (LS) \cite{xue2026multibench++} and TMC \cite{han2022trusted}, which use direct logit summation and uncertainty-aware evidential fusion, respectively.

\textbf{Model Training.} 
We train our model following the in-context learning paradigm, where each synthetic multimodal dataset is treated as an independent prediction task. During training, we perform approximately $12,000$ optimization steps, with each step sampling $16$ synthetic multimodal datasets, resulting in around $192,000$ unique synthetic tasks. The number of samples in each dataset is randomly generated within the range of $512$ to $2,048$, and each task is further divided into a context set and a query set, where the query size is fixed to $128$ while the remaining samples serve as contextual examples. We optimize the model using the Adam optimizer \cite{kingma2015adam} with a learning rate of $2\times10^{-4}$, employing a linear warm-up phase followed by cosine annealing for learning-rate scheduling \cite{loshchilov2017sgdr}.

\subsection{Main Results}
\begin{figure}[htbp]
  \begin{center}
    \centerline{\includegraphics[width=0.99\columnwidth]{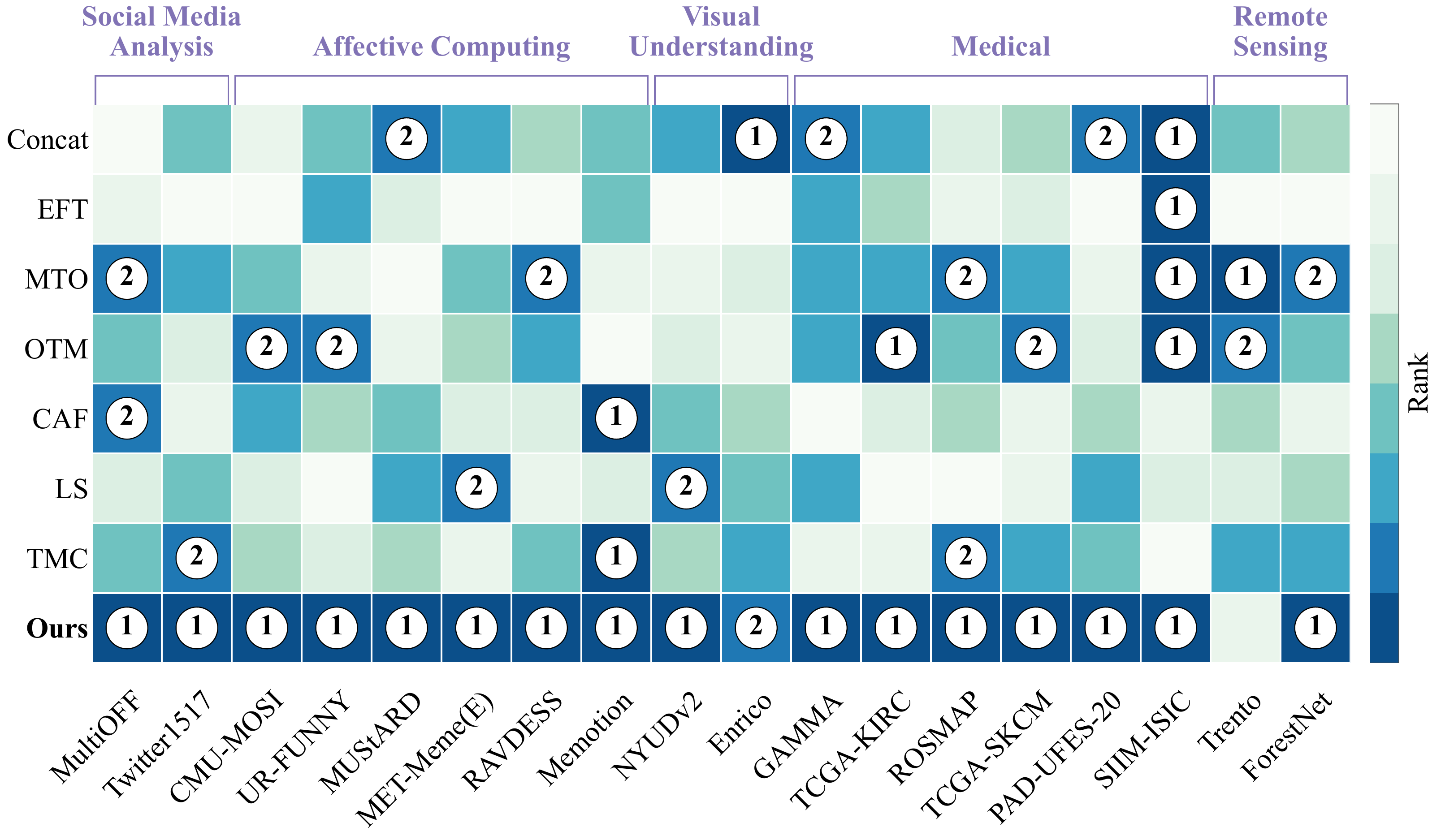}}
    \caption{Comparison of accuracy ranking between FAM and representative specialized fusion methods across diverse multimodal datasets. Our model achieves performance comparable to these specialized methods without parameter adaptation.}
    \label{rank}
  \end{center}
\end{figure}

Figure~\ref{rank} summarizes the accuracy rankings of the proposed FAM and representative specialized fusion methods across real-world multimodal datasets. No specialized fusion method consistently outperforms the others across all evaluated modality combinations and prediction tasks, with relative rankings varying substantially across datasets. In contrast, FAM achieves strong performance on nearly all evaluated datasets across diverse application domains. The tied or lower rankings of FAM on SIIM-ISIC and Trento reflect closely matched accuracy scores across methods, as shown in the detailed results in the Appendix B.1.

\begin{figure}[htbp]
  \centering
  \includegraphics[width=1.05\textwidth]{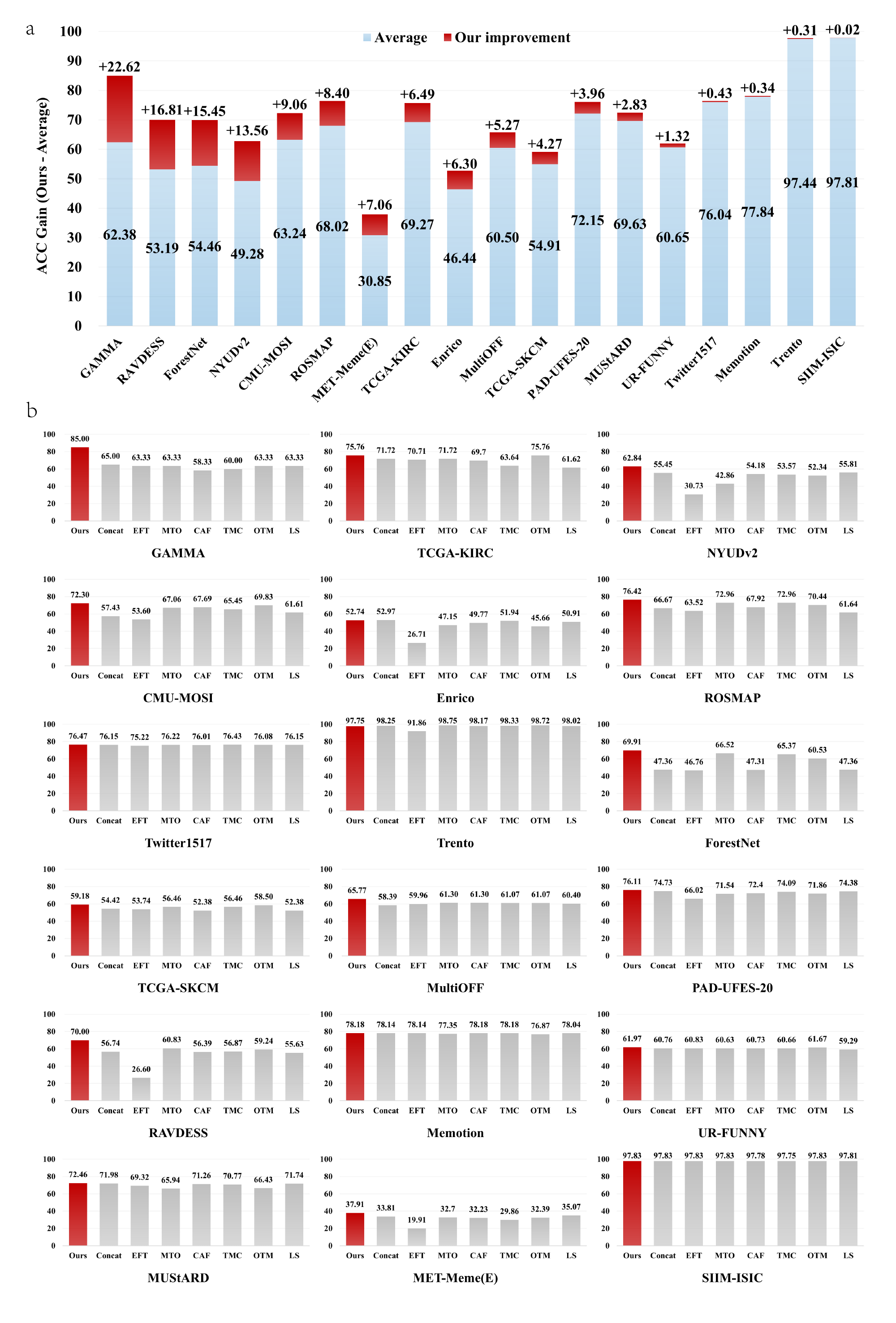}
  \caption{Comparison between FAM and specialized fusion models across diverse multimodal datasets.}
  \label{fig4}
\end{figure}

We further present detailed comparisons across the $18$ datasets reported in the main paper. As shown in Figure~\ref{fig4}, FAM achieves strong performance across diverse application domains, improving average ACC by \textbf{$6.92\%$} over the existing specialized fusion baselines. These results are obtained using our model without task-specific optimization or parameter updates during inference. Unlike specialized fusion models optimized separately for predefined modality combinations and datasets, FAM is trained on diverse synthetic multimodal tasks generated by SMCMs, encouraging it to learn transferable multimodal correlation patterns. During inference, labeled contextual examples from each real-world dataset provide task-specific information, allowing the model to draw on these learned patterns for prediction under the current modality combination and task setting. Overall, these results support the effectiveness of the paradigm of training on diverse synthetic multimodal datasets for generalization across real-world multimodal prediction tasks. For more comparison results, refer to Appendix.

\subsection{Ablation Study}

\textbf{Effectiveness of Multimodal Extension.} To evaluate the effectiveness of the proposed FAM in extending context-based prediction toward generalized multimodal scenarios, we compare it with TabPFN under the same input settings. Here, we first apply PCA to the raw modality representations and project them into the same token space, ensuring that both models receive the same number of input tokens for a fair comparison. As shown in Table~\ref{tab1}, our model exhibits improved overall performance across diverse datasets, indicating its enhanced capability in modeling multimodal correlations. These results demonstrate that training over diverse synthetic  datasets enables our Fusion Anything model to acquire richer multimodal correlation patterns and effectively extend the prediction capability of the base model to heterogeneous multimodal scenarios. Additional comparison results are provided in the Appendix B.2.

\begin{table}[htbp]
\centering
\caption{Comparison of the original TabPFN and the proposed FAM.}
\label{tab1}
\setlength{\tabcolsep}{2.7pt}
\resizebox{\textwidth}{!}{%
\begin{tabular}{lcccccccccccccccc}
\toprule
\multirow{2}{*}{Method}
& \multicolumn{2}{c}{GAMMA}
& \multicolumn{2}{c}{RAVDESS}
& \multicolumn{2}{c}{NYUDv2}
& \multicolumn{2}{c}{MET-Meme(E)}
& \multicolumn{2}{c}{TCGA-SKCM}
& \multicolumn{2}{c}{MultiOFF}
& \multicolumn{2}{c}{TCGA-KIRC}
& \multicolumn{2}{c}{ROSMAP} \\
\cmidrule(lr){2-3}\cmidrule(lr){4-5}\cmidrule(lr){6-7}\cmidrule(lr){8-9}\cmidrule(lr){10-11}\cmidrule(lr){12-13}\cmidrule(lr){14-15}\cmidrule(lr){16-17}
& ACC$\uparrow$ & AUC$\uparrow$
& ACC$\uparrow$ & AUC$\uparrow$
& ACC$\uparrow$ & AUC$\uparrow$
& ACC$\uparrow$ & AUC$\uparrow$
& ACC$\uparrow$ & AUC$\uparrow$
& ACC$\uparrow$ & AUC$\uparrow$
& ACC$\uparrow$ & AUC$\uparrow$
& ACC$\uparrow$ & AUC$\uparrow$ \\
\midrule
TabPFN
& 65.00 & 80.00
& 57.92 & 89.12
& 57.80 & 87.99
& 33.65 & 64.40
& 55.10 & 56.32
& 61.74 & 62.81
& 72.73 & \textbf{79.34}
& 73.58 & \textbf{86.35} \\
Ours
& \textbf{85.00} & \textbf{92.00}
& \textbf{70.00} & \textbf{94.25}
& \textbf{62.84} & \textbf{90.85}
& \textbf{37.91} & \textbf{67.02}
& \textbf{59.18} & \textbf{60.53}
& \textbf{65.77} & \textbf{68.44}
& \textbf{75.76} & 72.31
& \textbf{76.42} & 84.56 \\
\bottomrule
\end{tabular}%
}
\end{table}

\begin{figure}[htbp]
  \centering
  \includegraphics[width=0.9\linewidth]{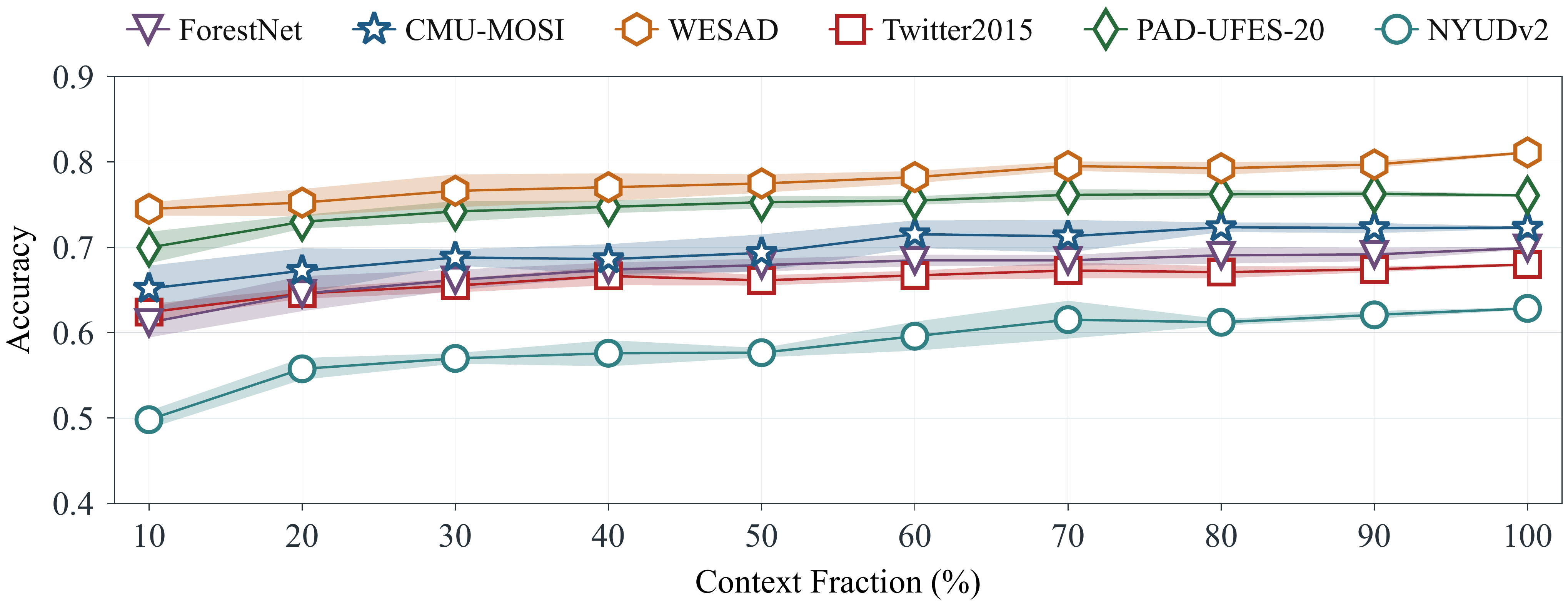}
    \caption{Performance scaling with different fractions of contextual examples.}
  \label{scaling}
\end{figure}

\textbf{Effect of Contextual Examples.}
To investigate the impact of contextual examples on prediction performance, we evaluate our Fusion Anything model under different context fractions while keeping the query set fixed. Specifically, contextual examples are randomly sampled from the available context pool with three random seeds $\{1,2,3\}$ at fractions ranging from $10\%$ to $100\%$. As shown in Figure~\ref{scaling}, our model exhibits an overall positive scaling trend across diverse multimodal datasets. The prediction performance generally improves as more contextual examples become available, while the marginal gains tend to diminish at larger context fractions. This trend indicates that Fusion Anything can effectively leverage contextual examples to identify task-relevant multimodal correlation patterns, while maintaining effective prediction even with limited contextual information. These results further demonstrate the stability of the model's in-context prediction capability across varying context sizes.

\textbf{Flexibility in the Choice of Pre-trained Encoders.} To demonstrate that the proposed FAM is not restricted to particular modality encoders, we evaluate its performance using different pre-trained encoder configurations. As reported in Table~\ref{tab2}, our model maintains strong prediction performance, achieving an average improvement of $8.55\%$ over the mean performance of baseline methods across all evaluated settings. This advantage is consistently observed across multiple datasets, suggesting that the learned multimodal correlations can be effectively utilized despite variations in the underlying feature representations. Although the quality of extracted representations inevitably affects the final performance, our model is able to activate and utilize appropriate correlation patterns learned during training, achieving comparable or superior results to baseline methods in most cases. These findings suggest that FAM learns transferable multimodal correlation patterns from synthetic datasets, enabling it to accommodate diverse modality representations without relying on a particular choice of encoders.

\begin{table}[htbp]
\centering
\caption{Comparison under different pre-trained encoder configurations. For each configuration, the best and second-best results among all methods are highlighted in bold and \underline{underlined}, respectively.}
\label{tab2}
\setlength{\tabcolsep}{3pt}
\resizebox{\linewidth}{!}{
\begin{tabular}{lllccccccc}
\toprule
Dataset & Modalities & Combination & Concat & EFT & MTO & CAF & TMC & Avg. & Ours \\
\midrule
\multirow{3}{*}{MultiOFF}
& \multirow{3}{*}{Image+Text}
& DINOv3+Qwen3
& 58.39 & 59.96 & \underline{61.30} & \underline{61.30} & 61.07 & 60.40 & \textbf{65.77} \\
& & ResNet18+Qwen3
& 58.61 & 57.27 & \underline{59.96} & 59.06 & 57.49 & 58.48 & \textbf{63.09} \\
& & DINOv3+BERT
& 62.42 & 60.40 & 56.82 & 59.73 & \underline{65.10} & 60.89 & \textbf{66.44} \\
\midrule
\multirow{3}{*}{RAVDESS}
& \multirow{3}{*}{Audio+Visual}
& WavLM+DINOv3
& 56.74 & 26.60 & \underline{60.83} & 56.39 & 56.87 & 51.49 & \textbf{70.00} \\
& & Whisper+DINOv3
& \underline{77.71} & 32.08 & 68.82 & 66.94 & \textbf{77.99} & 64.71 & 76.04 \\
& & WavLM+ResNet18
& 37.92 & 18.68 & \underline{47.08} & 37.64 & 35.90 & 35.44 & \textbf{58.54} \\
\midrule
\multirow{3}{*}{Twitter2015}
& \multirow{3}{*}{Image+Text}
& ResNet18+BERT
& 64.77 & 63.20 & 66.47 & \underline{67.12} & 64.51 & 65.21 & \textbf{67.98} \\
& & DINOv3+BERT
& 64.93 & 62.84 & \underline{65.57} & 64.87 & 64.13 & 64.47 & \textbf{67.31} \\
& & ResNet18+Qwen3
& 60.88 & 58.66 & 64.90 & \textbf{67.05} & 63.36 & 62.97 & \underline{65.86} \\
\bottomrule
\end{tabular}}
\end{table}

\section{Conclusion}
In this paper, we propose FAM, a generalized multimodal foundation model toward unified prediction beyond predefined modality combinations and prediction tasks. By constructing large-scale synthetic datasets with diverse causal structures, FAM learns transferable multimodal correlation patterns rather than task-specific fusion strategies. Through multimodal in-context learning, the model leverages contextual examples from unseen datasets to activate task-relevant correlation patterns and perform prediction without additional parameter updates. Extensive experiments demonstrate that our model achieves competitive or superior performance to state-of-the-art specialized fusion models across diverse multimodal benchmarks, while requiring no task-specific optimization at inference time. The proposed FAM provides a promising direction toward supporting future modalities beyond those explored by existing multimodal learning paradigms. Moreover, there are great potential in exploring applications (e.g., embodied intelligence, scientific research) in the future.

\bibliographystyle{unsrt} 
\bibliography{reference}

@article{han2022trusted,
  title={Trusted multi-view classification with dynamic evidential fusion},
  author={Han, Zongbo and Zhang, Changqing and Fu, Huazhu and Zhou, Joey Tianyi},
  journal={IEEE Transactions on Pattern Analysis and Machine Intelligence},
  volume={45},
  number={2},
  pages={2551--2566},
  year={2022}
}

@inproceedings{zhang2023provable,
  title={Provable dynamic fusion for low-quality multimodal data},
  author={Zhang, Qingyang and Wu, Haitao and Zhang, Changqing and Hu, Qinghua and Fu, Huazhu and Zhou, Joey Tianyi and Peng, Xi},
  booktitle={International Conference on Machine Learning},
  pages={41753--41769},
  year={2023}
}

@inproceedings{li2023blip,
  title={Blip-2: Bootstrapping language-image pre-training with frozen image encoders and large language models},
  author={Li, Junnan and Li, Dongxu and Savarese, Silvio and Hoi, Steven},
  booktitle={International conference on Machine Learning},
  pages={19730--19742},
  year={2023}
}

@article{wang2026multimodal,
  title={Multimodal learning with next-token prediction for large multimodal models},
  author={Wang, Xinlong and Cui, Yufeng and Wang, Jinsheng and Zhang, Fan and Wang, Yueze and Zhang, Xiaosong and Luo, Zhengxiong and Sun, Quan and Li, Zhen and Wang, Yuqi and others},
  journal={Nature},
  volume={650},
  number={8101},
  pages={327--333},
  year={2026}
}

@inproceedings{zhan2021product1m,
  title={Product1m: Towards weakly supervised instance-level product retrieval via cross-modal pretraining},
  author={Zhan, Xunlin and Wu, Yangxin and Dong, Xiao and Wei, Yunchao and Lu, Minlong and Zhang, Yichi and Xu, Hang and Liang, Xiaodan},
  booktitle={IEEE/CVF International Conference on Computer Vision},
  pages={11762--11771},
  year={2021}
}

@inproceedings{li2021ai,
  title={AI choreographer: Music conditioned 3d dance generation with aist++},
  author={Li, Ruilong and Yang, Shan and Ross, David A and Kanazawa, Angjoo},
  booktitle={IEEE/CVF International Conference on Computer Vision},
  pages={13381--13392},
  year={2021}
}

@article{liang2021multibench,
  title={Multibench: Multiscale benchmarks for multimodal representation learning},
  author={Liang, Paul Pu and Lyu, Yiwei and Fan, Xiang and Wu, Zetian and Cheng, Yun and Wu, Jason and Chen, Leslie and Wu, Peter and Lee, Michelle A and Zhu, Yuke and others},
  journal={Advances in Neural Information Processing Systems},
  volume={2021},
  number={DB1},
  pages={1},
  year={2021}
}

@inproceedings{xue2026multibench++,
  title={{MULTIBENCH++}: A Unified and Comprehensive Multimodal Fusion Benchmarking Across Specialized Domains},
  author={Xue, Leyan and Zhang, Changqing and Xue, Kecheng and Liu, Xiaohong and Wang, Guangyu and Han, Zongbo},
  booktitle={Proceedings of the AAAI Conference on Artificial Intelligence},
  volume={40},
  number={32},
  pages={27450--27458},
  year={2026}
}

@article{hollmann2025accurate,
  title={Accurate predictions on small data with a tabular foundation model},
  author={Hollmann, Noah and M{\"u}ller, Samuel and Purucker, Lennart and Krishnakumar, Arjun and K{\"o}rfer, Max and Hoo, Shi Bin and Schirrmeister, Robin Tibor and Hutter, Frank},
  journal={Nature},
  volume={637},
  number={8045},
  pages={319--326},
  year={2025}
}

@article{huang2026dissecting,
  title={Dissecting Multimodal In-Context Learning: Modality Asymmetries and Circuit Dynamics in modern Transformers},
  author={Huang, Yiran and Roth, Karsten and Bouniot, Quentin and Xu, Wenjia and Akata, Zeynep},
  journal={arXiv preprint arXiv:2601.20796},
  year={2026}
}

@article{zhang2018generalized,
  title={Generalized latent multi-view subspace clustering},
  author={Zhang, Changqing and Fu, Huazhu and Hu, Qinghua and Cao, Xiaochun and Xie, Yuan and Tao, Dacheng and Xu, Dong},
  journal={IEEE Transactions on Pattern Analysis and Machine Intelligence},
  volume={42},
  number={1},
  pages={86--99},
  year={2018}
}

@book{pearl2009causality,
  title={Causality},
  author={Pearl, Judea},
  year={2009},
  publisher={Cambridge university press}
}

@article{baltruvsaitis2018multimodal,
  title={Multimodal machine learning: A survey and taxonomy},
  author={Baltru{\v{s}}aitis, Tadas and Ahuja, Chaitanya and Morency, Louis-Philippe},
  journal={IEEE Transactions on Pattern Analysis and Machine Intelligence},
  volume={41},
  number={2},
  pages={423--443},
  year={2018}
}

@article{xu2023multimodal,
  title={Multimodal learning with transformers: A survey},
  author={Xu, Peng and Zhu, Xiatian and Clifton, David A},
  journal={IEEE Transactions on Pattern Analysis and Machine Intelligence},
  volume={45},
  number={10},
  pages={12113--12132},
  year={2023}
}

@article{hu2021lora,
  title={Lora: Low-rank adaptation of large language models},
  author={Hu, Edward J and Shen, Yelong and Wallis, Phillip and Allen-Zhu, Zeyuan and Li, Yuanzhi and Wang, Shean and Wang, Lu and Chen, Weizhu},
  journal={arXiv preprint arXiv:2106.09685},
  year={2021}
}

@article{wu2023gamma,
  title={Gamma challenge: glaucoma grading from multi-modality images},
  author={Wu, Junde and Fang, Huihui and Li, Fei and Fu, Huazhu and Lin, Fengbin and Li, Jiongcheng and Huang, Yue and Yu, Qinji and Song, Sifan and Xu, Xinxing and others},
  journal={Medical Image Analysis},
  volume={90},
  pages={102938},
  year={2023},
  publisher={Elsevier}
}

@inproceedings{niu2016sentiment,
  title={Sentiment analysis on multi-view social data},
  author={Niu, Teng and Zhu, Shiai and Pang, Lei and El Saddik, Abdulmotaleb},
  booktitle={International Conference on Multimedia Modeling},
  pages={15--27},
  year={2016},
  organization={Springer}
}

@article{rotemberg2021patient,
  title={A patient-centric dataset of images and metadata for identifying melanomas using clinical context},
  author={Rotemberg, Veronica and Kurtansky, Nicholas and Betz-Stablein, Brigid and Caffery, Liam and Chousakos, Emmanouil and Codella, Noel and Combalia, Marc and Dusza, Stephen and Guitera, Pascale and Gutman, David and others},
  journal={Scientific Data},
  volume={8},
  number={1},
  pages={34},
  year={2021},
  publisher={Nature Publishing Group UK London}
}

@inproceedings{silberman2012indoor,
  title={Indoor segmentation and support inference from rgbd images},
  author={Silberman, Nathan and Hoiem, Derek and Kohli, Pushmeet and Fergus, Rob},
  booktitle={European Conference on Computer Vision},
  pages={746--760},
  year={2012},
  organization={Springer}
}

@article{weinstein2013cancer,
  title={The cancer genome atlas pan-cancer analysis project},
  author={Weinstein, John N and Collisson, Eric A and Mills, Gordon B and Shaw, Kenna R and Ozenberger, Brad A and Ellrott, Kyle and Shmulevich, Ilya and Sander, Chris and Stuart, Joshua M},
  journal={Nature Genetics},
  volume={45},
  number={10},
  pages={1113--1120},
  year={2013},
  publisher={Nature Publishing Group}
}

@inproceedings{castro2019towards,
  title={Towards multimodal sarcasm detection (an obviously perfect paper)},
  author={Castro, Santiago and Hazarika, Devamanyu and P{\'e}rez-Rosas, Ver{\'o}nica and Zimmermann, Roger and Mihalcea, Rada and Poria, Soujanya},
  booktitle={Proceedings of the 57th Annual Meeting of the Association for Computational Linguistics},
  pages={4619--4629},
  year={2019}
}

@inproceedings{suryawanshi2020multimodal,
  title={Multimodal meme dataset (MultiOFF) for identifying offensive content in image and text},
  author={Suryawanshi, Shardul and Chakravarthi, Bharathi Raja and Arcan, Mihael and Buitelaar, Paul},
  booktitle={Proceedings of the second workshop on trolling, aggression and cyberbullying},
  pages={32--41},
  year={2020}
}

@article{livingstone2018ryerson,
  title={The Ryerson Audio-Visual Database of Emotional Speech and Song (RAVDESS): A dynamic, multimodal set of facial and vocal expressions in North American English},
  author={Livingstone, Steven R and Russo, Frank A},
  journal={PloS one},
  volume={13},
  number={5},
  pages={e0196391},
  year={2018},
  publisher={Public Library of Science San Francisco, CA USA}
}

@inproceedings{xu2022met,
  title={Met-meme: A multimodal meme dataset rich in metaphors},
  author={Xu, Bo and Li, Tingting and Zheng, Junzhe and Naseriparsa, Mehdi and Zhao, Zhehuan and Lin, Hongfei and Xia, Feng},
  booktitle={Proceedings of the 45th International ACM SIGIR Conference on Research and Development in Information Retrieval},
  pages={2887--2899},
  year={2022}
}

@inproceedings{poria2019meld,
  title={Meld: A multimodal multi-party dataset for emotion recognition in conversations},
  author={Poria, Soujanya and Hazarika, Devamanyu and Majumder, Navonil and Naik, Gautam and Cambria, Erik and Mihalcea, Rada},
  booktitle={Proceedings of the 57th Annual Meeting of the Association for Computational Linguistics},
  pages={527--536},
  year={2019}
}

@inproceedings{schmidt2018introducing,
  title={Introducing {wesad}, a multimodal dataset for wearable stress and affect detection},
  author={Schmidt, Philip and Reiss, Attila and Duerichen, Robert and Marberger, Claus and Van Laerhoven, Kristof},
  booktitle={Proceedings of the 20th ACM International Conference on Multimodal Interaction},
  pages={400--408},
  year={2018}
}

@article{alayrac2022flamingo,
  title={Flamingo: a visual language model for few-shot learning},
  author={Alayrac, Jean-Baptiste and Donahue, Jeff and Luc, Pauline and Miech, Antoine and Barr, Iain and Hasson, Yana and Lenc, Karel and Mensch, Arthur and Millican, Katherine and Reynolds, Malcolm and others},
  journal={Advances in Neural Information Processing Systems},
  volume={35},
  pages={23716--23736},
  year={2022}
}

@article{liang2024foundations,
  title={Foundations \& trends in multimodal machine learning: Principles, challenges, and open questions},
  author={Liang, Paul Pu and Zadeh, Amir and Morency, Louis-Philippe},
  journal={ACM Computing Surveys},
  volume={56},
  number={10},
  pages={1--42},
  year={2024},
  publisher={ACM New York, NY}
}

@inproceedings{wei2020multi,
  title={Multi-modality cross attention network for image and sentence matching},
  author={Wei, Xi and Zhang, Tianzhu and Li, Yan and Zhang, Yongdong and Wu, Feng},
  booktitle={IEEE/CVF Conference on Computer Vision and Pattern Recognition},
  pages={10938--10947},
  year={2020}
}

@inproceedings{sharma2020semeval,
  title={SemEval-2020 task 8: Memotion analysis-the visuo-lingual metaphor!},
  author={Sharma, Chhavi and Bhageria, Deepesh and Scott, William and Pykl, Srinivas and Das, Amitava and Chakraborty, Tanmoy and Pulabaigari, Viswanath and Gamb{\"a}ck, Bj{\"o}rn},
  booktitle={Proceedings of the Workshop on Semantic Evaluation},
  pages={759--773},
  year={2020}
}

@article{ramachandram2017deep,
  title={Deep multimodal learning: A survey on recent advances and trends},
  author={Ramachandram, Dhanesh and Taylor, Graham W},
  journal={IEEE Signal Processing Magazine},
  volume={34},
  number={6},
  pages={96--108},
  year={2017},
  publisher={IEEE}
}

@article{atrey2010multimodal,
  title={Multimodal fusion for multimedia analysis: a survey},
  author={Atrey, Pradeep K and Hossain, M Anwar and El Saddik, Abdulmotaleb and Kankanhalli, Mohan S},
  journal={Multimedia Systems},
  volume={16},
  number={6},
  pages={345--379},
  year={2010},
  publisher={Springer}
}

@article{nagrani2021attention,
  title={Attention bottlenecks for multimodal fusion},
  author={Nagrani, Arsha and Yang, Shan and Arnab, Anurag and Jansen, Aren and Schmid, Cordelia and Sun, Chen},
  journal={Advances in Neural Information Processing Systems},
  volume={34},
  pages={14200--14213},
  year={2021}
}

@inproceedings{girdhar2022omnivore,
  title={Omnivore: A single model for many visual modalities},
  author={Girdhar, Rohit and Singh, Mannat and Ravi, Nikhila and Van Der Maaten, Laurens and Joulin, Armand and Misra, Ishan},
  booktitle={IEEE/CVF Conference on Computer Vision and Pattern Recognition},
  pages={16081--16091},
  year={2022},
  organization={IEEE}
}

@inproceedings{wang2022multimodal,
  title={Multimodal token fusion for vision transformers},
  author={Wang, Yikai and Chen, Xinghao and Cao, Lele and Huang, Wenbing and Sun, Fuchun and Wang, Yunhe},
  booktitle={IEEE/CVF Conference on Computer Vision and Pattern Recognition},
  pages={12176--12185},
  year={2022},
  organization={IEEE}
}

@inproceedings{sun2021multimodal,
  title={Multimodal cross-and self-attention network for speech emotion recognition},
  author={Sun, Licai and Liu, Bin and Tao, Jianhua and Lian, Zheng},
  booktitle={IEEE International Conference on Acoustics, Speech and Signal Processing},
  pages={4275--4279},
  year={2021},
  organization={IEEE}
}

@inproceedings{radford2021learning,
  title={Learning transferable visual models from natural language supervision},
  author={Radford, Alec and Kim, Jong Wook and Hallacy, Chris and Ramesh, Aditya and Goh, Gabriel and Agarwal, Sandhini and Sastry, Girish and Askell, Amanda and Mishkin, Pamela and Clark, Jack and others},
  booktitle={International Conference on Machine Learning},
  pages={8748--8763},
  year={2021},
  organization={PmLR}
}

@inproceedings{jia2021scaling,
  title={Scaling up visual and vision-language representation learning with noisy text supervision},
  author={Jia, Chao and Yang, Yinfei and Xia, Ye and Chen, Yi-Ting and Parekh, Zarana and Pham, Hieu and Le, Quoc and Sung, Yun-Hsuan and Li, Zhen and Duerig, Tom},
  booktitle={International Conference on Machine Learning},
  pages={4904--4916},
  year={2021},
  organization={PMLR}
}

@article{team2024chameleon,
  title={Chameleon: Mixed-modal early-fusion foundation models, 2024},
  author={Team, Chameleon},
  journal={URL https://arxiv. org/abs/2405.09818},
  volume={9},
  number={8},
  year={2024}
}

@inproceedings{wu2025janus,
  title={Janus: Decoupling visual encoding for unified multimodal understanding and generation},
  author={Wu, Chengyue and Chen, Xiaokang and Wu, Zhiyu and Ma, Yiyang and Liu, Xingchao and Pan, Zizheng and Liu, Wen and Xie, Zhenda and Yu, Xingkai and Ruan, Chong and others},
  booktitle={IEEE/CVF Conference on Computer Vision and Pattern Recognition},
  pages={12966--12977},
  year={2025},
  organization={IEEE}
}

@article{liu2023visual,
  title={Visual instruction tuning},
  author={Liu, Haotian and Li, Chunyuan and Wu, Qingyang and Lee, Yong Jae},
  journal={Advances in Neural Information Processing Systems},
  volume={36},
  pages={34892--34916},
  year={2023}
}

@article{dong2026advances,
  title={Advances in multimodal adaptation and generalization: From traditional approaches to foundation models},
  author={Dong, Hao and Liu, Moru and Zhou, Kaiyang and Chatzi, Eleni and Kannala, Juho and Stachniss, Cyrill and Fink, Olga},
  journal={IEEE Transactions on Pattern Analysis and Machine Intelligence},
  year={2026},
  publisher={IEEE}
}

@article{paige2017learning,
  title={Learning disentangled representations with semi-supervised deep generative models},
  author={Paige, Brooks and Van De Meent, Jan-Willem and Desmaison, Alban and Goodman, Noah and Kohli, Pushmeet and Wood, Frank and Torr, Philip and others},
  journal={Advances in Neural Information Processing Systems},
  volume={30},
  year={2017}
}

@article{zhang2026multimodal,
  title={Multimodal fusion on low-quality data: A comprehensive survey},
  author={Zhang, Qingyang and Wei, Yake and Han, Zongbo and Fu, Huazhu and Peng, Xi and Hu, Qinghua and Deng, Cheng and Xu, Cai and Wen, Jie and Hu, Di and others},
  journal={Information Fusion},
  pages={104437},
  year={2026},
  publisher={Elsevier}
}

@inproceedings{leiva2020enrico,
  title={Enrico: A dataset for topic modeling of mobile UI designs},
  author={Leiva, Luis A and Hota, Asutosh and Oulasvirta, Antti},
  booktitle={International Conference on Human-Computer Interaction with Mobile Devices and Services},
  pages={1--4},
  year={2020}
}

@article{pacheco2020pad,
  title={PAD-UFES-20: A skin lesion dataset composed of patient data and clinical images collected from smartphones},
  author={Pacheco, Andre GC and Lima, Gustavo R and Salomao, Amanda S and Krohling, Breno and Biral, Igor P and De Angelo, Gabriel G and Alves Jr, F{\'a}bio CR and Esgario, Jos{\'e} GM and Simora, Alana C and Castro, Pedro BC and others},
  journal={Data in Brief},
  volume={32},
  pages={106221},
  year={2020},
  publisher={Elsevier}
}

@article{liang2023quantifying,
  title={Quantifying \& modeling multimodal interactions: An information decomposition framework},
  author={Liang, Paul Pu and Cheng, Yun and Fan, Xiang and Ling, Chun Kai and Nie, Suzanne and Chen, Richard and Deng, Zihao and Allen, Nicholas and Auerbach, Randy and Mahmood, Faisal and others},
  journal={Advances in Neural Information Processing Systems},
  volume={36},
  pages={27351--27393},
  year={2023}
}

@article{bommasani2021opportunities,
  title={On the opportunities and risks of foundation models},
  author={Bommasani, Rishi and Hudson, Drew A and Adeli, Ehsan and Altman, Russ and Arora, Simran and von Arx, Sydney and Bernstein, Michael S and Bohg, Jeannette and Bosselut, Antoine and Brunskill, Emma and others},
  journal={arXiv preprint arXiv:2108.07258},
  year={2021}
}

@article{haque2020illuminating,
  title={Illuminating the dark spaces of healthcare with ambient intelligence},
  author={Haque, Albert and Milstein, Arnold and Fei-Fei, Li},
  journal={Nature},
  volume={585},
  number={7824},
  pages={193--202},
  year={2020},
  publisher={Nature Publishing Group UK London}
}

@article{wang2026limix,
  title={LimiX-2M: Mitigating Low-Rank Collapse and Attention Bottlenecks in Tabular Foundation Models},
  author={Wang, Yuanrui and Zhang, Xingxuan and Yu, Han and Hao, Mingchao and Ren, Gang and Yuan, Hao and Mao, Li and Zhang, Yunjia and Yuan, Chun and Cui, Peng},
  journal={arXiv preprint arXiv:2606.04485},
  year={2026}
}

@inproceedings{kirillov2023segment,
  title={Segment anything},
  author={Kirillov, Alexander and Mintun, Eric and Ravi, Nikhila and Mao, Hanzi and Rolland, Chloe and Gustafson, Laura and Xiao, Tete and Whitehead, Spencer and Berg, Alexander C and Lo, Wan-Yen and others},
  booktitle={2023 IEEE/CVF International Conference on Computer Vision (ICCV)},
  pages={3992--4003},
  year={2023},
  organization={IEEE}
}

@article{kim2026multimodalpfn,
  title={Multimodalpfn: Extending prior-data fitted networks for multimodal tabular learning},
  author={Kim, Wall and Song, Chaeyoung and Kim, Hanul},
  journal={arXiv preprint arXiv:2602.20223},
  year={2026}
}

@article{zhang2019cpm,
  title={{CPM-Nets}: Cross partial multi-view networks},
  author={Zhang, Changqing and Han, Zongbo and Fu, Huazhu and Zhou, Joey Tianyi and Hu, Qinghua and others},
  journal={Advances in Neural Information Processing Systems},
  volume={32},
  year={2019}
}

@inproceedings{han2022multimodal,
  title={Multimodal dynamics: Dynamical fusion for trustworthy multimodal classification},
  author={Han, Zongbo and Yang, Fan and Huang, Junzhou and Zhang, Changqing and Yao, Jianhua},
  booktitle={IEEE/CVF Conference on Computer Vision and Pattern Recognition},
  pages={20675--20685},
  year={2022},
  organization={IEEE}
}

@inproceedings{yang2024depth,
  title={Depth Anything: Unleashing the power of large-scale unlabeled data},
  author={Yang, Lihe and Kang, Bingyi and Huang, Zilong and Xu, Xiaogang and Feng, Jiashi and Zhao, Hengshuang},
  booktitle={2024 IEEE/CVF Conference on Computer Vision and Pattern Recognition (CVPR)},
  pages={10371--10381},
  year={2024},
  organization={IEEE}
}

@article{lee2026mitigating,
  title={Mitigating Label Shift in Tabular In-Context Learning via Test-Time Posterior Adjustment},
  author={Lee, Seunghan},
  journal={arXiv preprint arXiv:2605.04363},
  year={2026}
}

@article{ma2026tabdpt,
  title={Tabdpt: Scaling tabular foundation models on real data},
  author={Ma, Junwei and Thomas, Valentin and Hosseinzadeh, Rasa and Labach, Alex and Cresswell, Jesse and Golestan, Keyvan and Yu, Guangwei and Caterini, Anthony L and Volkovs, Maks},
  journal={Advances in Neural Information Processing Systems},
  volume={38},
  pages={172692--172722},
  year={2026}
}

@article{zadeh2016mosi,
  title={Mosi: multimodal corpus of sentiment intensity and subjectivity analysis in online opinion videos},
  author={Zadeh, Amir and Zellers, Rowan and Pincus, Eli and Morency, Louis-Philippe},
  journal={arXiv preprint arXiv:1606.06259},
  year={2016}
}

@article{bennett2018religious,
  title={Religious orders study and rush memory and aging project},
  author={Bennett, David A and Buchman, Aron S and Boyle, Patricia A and Barnes, Lisa L and Wilson, Robert S and Schneider, Julie A},
  journal={Journal of Alzheimer’s disease},
  volume={64},
  number={s1},
  pages={S161--S189},
  year={2018},
  publisher={SAGE Publications Sage UK: London, England}
}

@inproceedings{hasan2019ur,
  title={UR-FUNNY: A multimodal language dataset for understanding humor},
  author={Hasan, Md Kamrul and Rahman, Wasifur and Zadeh, AmirAli Bagher and Zhong, Jianyuan and Tanveer, Md Iftekhar and Morency, Louis-Philippe and Hoque, Mohammed Ehsan},
  booktitle={Proceedings of the 2019 conference on empirical methods in natural language processing and the 9th international joint conference on natural language processing (EMNLP-IJCNLP)},
  pages={2046--2056},
  year={2019}
}

@article{irvin2020forestnet,
  title={Forestnet: Classifying drivers of deforestation in indonesia using deep learning on satellite imagery},
  author={Irvin, Jeremy and Sheng, Hao and Ramachandran, Neel and Johnson-Yu, Sonja and Zhou, Sharon and Story, Kyle and Rustowicz, Rose and Elsworth, Cooper and Austin, Kemen and Ng, Andrew Y},
  journal={arXiv preprint arXiv:2011.05479},
  year={2020}
}

@inproceedings{song2015sun,
  title={{Sun RGB-D}: A rgb-d scene understanding benchmark suite},
  author={Song, Shuran and Lichtenberg, Samuel P and Xiao, Jianxiong},
  booktitle={Proceedings of the IEEE conference on Computer Vision and Pattern Recognition},
  pages={567--576},
  year={2015}
}

@article{lu2019vilbert,
  title={Vilbert: Pretraining task-agnostic visiolinguistic representations for vision-and-language tasks},
  author={Lu, Jiasen and Batra, Dhruv and Parikh, Devi and Lee, Stefan},
  journal={Advances in Neural Information Processing Systems},
  volume={32},
  year={2019}
}

@article{lin2020interbert,
  title={Interbert: Vision-and-language interaction for multi-modal pretraining},
  author={Lin, Junyang and Yang, An and Zhang, Yichang and Liu, Jie and Zhou, Jingren and Yang, Hongxia},
  journal={arXiv preprint arXiv:2003.13198},
  year={2020}
}

@article{kingma2015adam,
  title={Adam: A method for stochastic optimization},
  author={Kingma, Diederik P and Ba, Jimmy},
  journal={International Conference on Learning Representations},
  year={2015}
}

@article{loshchilov2017sgdr,
  title={Sgdr: Stochastic gradient descent with warm restarts},
  author={Loshchilov, Ilya and Hutter, Frank},
  journal={International Conference on Learning Representations},
  year={2017}
}


\clearpage
\appendix
\section{Dataset Details}

\subsection*{GAMMA}
\noindent\textbf{Data Source and Content} GAMMA \cite{wu2023gamma} is from the Glaucoma grAding from Multi-Modality imAges challenge. It contains color fundus images and OCT volumes for three-class glaucoma grading.

\noindent\textbf{Link} \url{https://zenodo.org/records/15119049}

\noindent\textbf{Feature Extraction} Both the fundus images and OCT volumes are encoded using an ImageNet-pretrained ResNet-18.

\noindent\textbf{Data Splits} A stratified split with seed 42 first assigns 80 samples to the non-test partition and 20 samples to the test set. The non-test partition is then divided using another stratified split with seed 10, yielding 64 training, 16 validation, and 20 test samples.

\subsection*{TCGA-KIRC}
\noindent\textbf{Data Source and Content} TCGA-KIRC \cite{weinstein2013cancer} contains aligned molecular measurements for kidney renal clear cell carcinoma patients. Six omics views are used for binary survival-status classification.

\noindent\textbf{Link} \url{https://portal.gdc.cancer.gov/projects/TCGA-KIRC}

\noindent\textbf{Feature Extraction} CNV, DNA methylation, miRNA, RNA-seq, RPPA, and cell-composition measurements are used directly. Each omics modality is represented using an identity mapping encoder.

\noindent\textbf{Data Splits} A fixed non-stratified split with seed 10 is used, containing 92 training, 41 validation, and 33 test patients.

\subsection*{NYUDv2}
\noindent\textbf{Data Source and Content} The NYU-Depth Dataset V2 (NYUDv2) \cite{silberman2012indoor} provides RGB and depth images of various indoor scenes captured using a Microsoft Kinect. We use 10 scene categories for classification.

\noindent\textbf{Link} \url{https://cs.nyu.edu/~fergus/datasets/nyu_depth_v2.html}

\noindent\textbf{Feature Extraction} An ImageNet-pretrained ResNet-18 is used for the RGB images and another for the depth images.

\noindent\textbf{Data Splits} The dataset is split into 381 training, 414 validation, and 654 test samples.

\subsection*{CMU-MOSI}
\noindent\textbf{Data Source and Content} CMU-MOSI \cite{zadeh2016mosi} contains opinion video segments with aligned visual, acoustic, and language signals and sentiment-intensity annotations. We use it for binary positive versus non-positive sentiment classification.

\noindent\textbf{Link} \url{https://github.com/CMU-MultiComp-Lab/CMU-MultimodalSDK}

\noindent\textbf{Feature Extraction} We use word-aligned FACET 4.2 visual features, COVAREP acoustic features, and GloVe 840B language features. After removing left padding using the language mask, each modality is averaged over its valid temporal positions.

\noindent\textbf{Data Splits} The official split is used, containing 1,283 training, 214 validation, and 686 test segments.

\subsection*{Enrico}
\noindent\textbf{Data Source and Content} Enrico \cite{leiva2020enrico} contains mobile UI screenshots and corresponding semantic wireframes for interface-topic classification. The original 20 design topics are merged into ten functional classes.

\noindent\textbf{Link} \url{https://github.com/luileito/enrico}

\noindent\textbf{Feature Extraction} The screenshots and semantic wireframes are independently encoded using an ImageNet-pretrained ResNet-18.

\noindent\textbf{Data Splits} A non-stratified split with seed 0 first assigns 80\% of the samples to the development partition and 20\% to the test set. The development partition is then divided using another non-stratified split with seed 0, yielding 992 training, 176 validation, and 292 test interfaces.

\subsection*{ROSMAP}
\noindent\textbf{Data Source and Content} ROSMAP \cite{bennett2018religious}contains multi-omics data from post-mortem brain tissue collected by the Religious Orders Study and Memory and Aging Project. We use it for binary classification of Alzheimer's disease and normal controls.

\noindent\textbf{Link} \url{https://github.com/txWang/MOGONET}

\noindent\textbf{Feature Extraction} Each tabular omics modality is encoded using an independent identity mapping encoder.

\noindent\textbf{Data Splits} The released split first assigns 245 samples to the non-test partition and 106 samples to the test set. The non-test partition is then divided using a stratified split with seed 10, yielding 196 training, 49 validation, and 106 test samples.

\subsection*{UR-FUNNY}
\noindent\textbf{Data Source and Content} UR-FUNNY \cite{hasan2019ur} contains aligned language, acoustic, and visual observations from public speeches for binary multimodal humor detection.

\noindent\textbf{Link} \url{https://github.com/ROC-HCI/UR-FUNNY}

\noindent\textbf{Feature Extraction} We use the released GloVe language embeddings, COVAREP acoustic features, and OpenFace visual features. The specialized models use temporal means, while the PCA48 TabPFN pipeline concatenates the temporal mean and standard deviation of each modality.

\noindent\textbf{Data Splits} The official split is used, containing 7,614 training, 980 validation, and 994 test samples.

\subsection*{TCGA-SKCM}
\noindent\textbf{Data Source and Content} TCGA-SKCM \cite{weinstein2013cancer} contains aligned molecular measurements for skin cutaneous melanoma patients. Six omics views from complete-case patients are used for binary survival-status classification.

\noindent\textbf{Link} \url{https://portal.gdc.cancer.gov/projects/TCGA-SKCM}

\noindent\textbf{Feature Extraction} CNV, DNA methylation, miRNA, RNA-seq, RPPA, and cell-composition measurements are used directly. Each omics modality is represented using an identity mapping encoder.

\noindent\textbf{Data Splits} A fixed non-stratified split with seed 10 is used, containing 137 training, 61 validation, and 49 test patients.

\subsection*{MultiOFF}
\noindent\textbf{Data Source and Content} MultiOFF \cite{suryawanshi2020multimodal} contains image--text memes for binary offensive-content detection.

\noindent\textbf{Link} \url{https://github.com/bharathichezhiyan/Multimodal-Meme-Classification-Identifying-Offensive-Content-in-Image-and-Text}

\noindent\textbf{Feature Extraction} Images are encoded using a frozen DINOv3 ViT-B/16, and text is encoded using Qwen3-Embedding-0.6B.

\noindent\textbf{Data Splits} The official split is used, containing 445 training, 149 validation, and 149 test posts.

\subsection*{MET-Meme(E)}
\noindent\textbf{Data Source and Content} The English subset of MET-Meme  \cite{xu2022met} contains image--text memes for seven-class sentiment classification.

\noindent\textbf{Link} \url{https://github.com/liaolianfoka/MET-Meme-A-Multi-modal-Meme-Dataset-Rich-in-Metaphors}

\noindent\textbf{Feature Extraction} Meme images are encoded using an ImageNet-pretrained ResNet-18, and text is encoded using BERT-base-uncased with attention-mask mean pooling.

\noindent\textbf{Data Splits} A non-stratified 7:1:2 random split with seed 42 is used, containing 737 training, 105 validation, and 211 test samples.

\subsection*{RAVDESS}
\noindent\textbf{Data Source and Content} The Ryerson Audio-Visual Database of Emotional Speech and Song (RAVDESS) \cite{livingstone2018ryerson} contains recordings from 24 professional actors. We use its speech subset and pair each official audio-only recording with the corresponding video-only recording for eight-class emotion recognition: neutral, calm, happy, sad, angry, fearful, disgust, and surprised.

\noindent\textbf{Link} \url{https://zenodo.org/records/1188976}

\noindent\textbf{Feature Extraction} Audio recordings are encoded using WavLM Base+, while eight uniformly sampled video frames are encoded using DINOv3 ViT-B/16 and averaged to obtain visual representations.

\noindent\textbf{Data Splits} We use a fixed actor-disjoint split. Actors 4, 10, 16, and 22 form the 240 validation utterances; actors 5, 6, 11, 12, 17, 18, 23, and 24 form the 480 test utterances; and the remaining actors form the 720 training utterances.

\subsection*{Memotion}
\noindent\textbf{Data Source and Content} Memotion \cite{sharma2020semeval} contains memes with transcribed text and affective annotations. We use its binary humor task, treating not-funny as the negative class and merging funny, very funny, and hilarious into the positive class.

\noindent\textbf{Link} \url{https://www.kaggle.com/datasets/williamscott701/memotion-dataset-7k}

\noindent\textbf{Feature Extraction} Meme images are encoded using DINOv3 ViT-B/16, while transcribed text is encoded using Qwen3-Embedding-0.6B.

\noindent\textbf{Data Splits} The processed dataset contains 6,831 samples. A fixed non-stratified 8:1:1 split with seed 42 is used, containing 5,465 training, 683 validation, and 683 test samples.

\subsection*{Twitter1517}
\noindent\textbf{Data Source and Content} Twitter1517 \cite{xue2026multibench++} contains paired tweet text and images for binary image--text correlation classification.

\noindent\textbf{Link} \url{https://github.com/code-chendl/HFIR}

\noindent\textbf{Feature Extraction} Images are encoded using a frozen DINOv3 ViT-B/16, and text is encoded using Qwen3-Embedding-0.6B.

\noindent\textbf{Data Splits} A stratified 7:1:2 split with seed 42 is used, containing 3,270 training, 467 validation, and 935 test samples.

\subsection*{Trento}
\noindent\textbf{Data Source and Content} Trento  \cite{xue2026multibench++} covers a rural area south of Trento, Italy. It contains co-registered hyperspectral imagery and a LiDAR-derived digital surface model for six-class land-cover classification.

\noindent\textbf{Link} \url{https://github.com/tyust-dayu/Trento/tree/b4afc449ce5d6936ddc04fe267d86f9f35536afd}

\noindent\textbf{Feature Extraction} A frozen HyperFree-b model encodes $5\times5$ hyperspectral patches, while a frozen DOFA ViT-Base model encodes the corresponding $5\times5$ digital-surface-model patches.

\noindent\textbf{Data Splits} A stratified 7:1:2 split with seed 42 first defines 24,170 non-test and 6,044 test samples. The non-test partition is then divided using a stratified 8:2 split with seed 10, yielding 19,336 training, 4,834 validation, and 6,044 test samples.

\subsection*{ForestNet}
\noindent\textbf{Data Source and Content} ForestNet \cite{irvin2020forestnet} contains satellite observations of forest-loss events in Indonesia together with expert annotations of their direct drivers. We use the four merged classes: plantation, smallholder agriculture, grassland/shrubland, and other.

\noindent\textbf{Link} \url{https://stanfordmlgroup.github.io/projects/forestnet/}

\noindent\textbf{Feature Extraction} RGB imagery and three-channel SRTM data are encoded using a shared frozen ImageNet-pretrained ResNet-152. Latitude, longitude, and year are retained as a three-dimensional structured modality using identity mapping.

\noindent\textbf{Data Splits} The processed official split is retained without additional randomization, containing 1,616 training, 473 validation, and 668 test samples.

\subsection*{MUStARD}
\noindent\textbf{Data Source and Content} MUStARD \cite{castro2019towards}contains 690 aligned text, audio, and video utterances from television shows for binary sarcasm detection.

\noindent\textbf{Link} \url{https://github.com/soujanyaporia/MUStARD}

\noindent\textbf{Feature Extraction} Text is represented by the mean of the last four BERT-base-uncased CLS layers. The officially released 283-dimensional librosa audio features are averaged over temporal bins, and ImageNet-pretrained ResNet-152 pool5 video features are averaged over frames.

\noindent\textbf{Data Splits} Fold 1 of the official speaker-dependent stratified five-fold split is used. Its 552 non-test utterances are divided using a stratified 8:2 split with seed 10, yielding 441 training, 111 validation, and 138 test utterances.

\subsection*{PAD-UFES-20}
\noindent\textbf{Data Source and Content} PAD-UFES-20 \cite{pacheco2020pad} contains 2,298 smartphone clinical images from 1,373 patients, together with patient metadata for six-class skin-lesion diagnosis.

\noindent\textbf{Link} \url{https://data.mendeley.com/datasets/zr7vgbcyr2}

\noindent\textbf{Feature Extraction} Clinical images are encoded using a frozen ImageNet-1K V2 ResNet-50. Patient metadata is retained as a structured modality after missing-value handling and categorical encoding fitted on each outer non-test partition.

\noindent\textbf{Data Splits} A patient-grouped stratified five-fold split with seed 42 is used. Within each outer non-test partition, a patient-grouped stratified five-fold split with seed 10 defines the validation set, yielding 1,470--1,471 training, 368 validation, and 459--460 test samples per fold. Results are averaged over the five outer folds.

\subsection*{SIIM-ISIC}
\noindent\textbf{Data Source and Content} SIIM-ISIC \cite{rotemberg2021patient} is from the 2020 SIIM-ISIC Melanoma Classification challenge. It contains dermoscopic images of skin lesions and patient-level metadata for binary melanoma classification.

\noindent\textbf{Link} \url{https://www.kaggle.com/competitions/siim-isic-melanoma-classification/data}

\noindent\textbf{Feature Extraction} Dermoscopic images are encoded using an ImageNet-pretrained ResNet-18. Sex, approximate age, and anatomical site are retained as a structured modality using identity mapping after missing-value handling and categorical encoding fitted on the non-test partition.

\noindent\textbf{Data Splits} A fixed non-stratified 8:1:1 split with seed 42 is used, containing 26,502 training, 3,312 validation, and 3,312 test samples.

\subsection*{MELD}
\noindent\textbf{Data Source and Content} The Multimodal EmotionLines Dataset (MELD) \cite{poria2019meld} contains 13,708 utterances from multi-party conversations in the television series Friends, with aligned text, audio, and video. We use the three-class sentiment-classification task.

\noindent\textbf{Link} \url{https://github.com/declare-lab/MELD}

\noindent\textbf{Feature Extraction} We use the MultiBench-released utterance-level GloVe-average text features and feature-selected acoustic features directly through identity mapping encoders.

\noindent\textbf{Data Splits} The MultiBench split is used, containing 9,989 training, 1,109 validation, and 2,610 test utterances.

\subsection*{WESAD}
\noindent\textbf{Data Source and Content} WESAD  \cite{schmidt2018introducing} is a wearable stress-and-affect dataset recorded from 15 subjects using chest- and wrist-worn sensors. We retain the baseline, stress, and amusement states for three-class classification.

\noindent\textbf{Link} \url{https://archive.ics.uci.edu/dataset/465/wesad}

\noindent\textbf{Feature Extraction} Chest channels are jointly encoded using a frozen MOMENT-1-small, while the four wrist signals are encoded separately and averaged. Embeddings from consecutive 512-sample blocks are averaged within each 60-second window.

\noindent\textbf{Data Splits} Each contiguous retained-state interval is divided into non-overlapping 60-second windows, yielding 535 windows. A deterministic 15-fold leave-one-subject-out split is used. In each fold, one subject forms the test set, another forms the validation set according to a balanced schedule with seed 10, and the remaining 13 form the training set, yielding 462--465 training, 35--37 validation, and 35--37 test windows. Results are averaged over the 15 folds.

\subsection*{Twitter2015}
\noindent\textbf{Data Source and Content} Twitter2015 \cite{xue2026multibench++} originates from SemEval tasks and pairs tweets with relevant images for three-class multimodal aspect-based sentiment analysis.

\noindent\textbf{Link} \url{https://archive.org/details/twitterstream}

\noindent\textbf{Feature Extraction} Images are encoded using a frozen ImageNet-pretrained ResNet-18. The two text fields are concatenated and encoded using BERT-base-uncased with attention-mask mean pooling.

\noindent\textbf{Data Splits} The official split is used, containing 3,179 training, 1,122 validation, and 1,037 test samples.

\subsection*{SUN RGB-D}
\noindent\textbf{Data Source and Content} SUN RGB-D \cite{song2015sun} is a large-scale indoor-scene benchmark containing paired RGB and depth images captured using multiple RGB-D sensors. We use the fixed 19-category subset and merge the original categories into eight semantic classes: bathroom, residential, education, workspace, meeting, dining, service, and furniture.

\noindent\textbf{Link} \url{https://rgbd.cs.princeton.edu/}

\noindent\textbf{Feature Extraction} The RGB images and depth maps are independently encoded using an ImageNet-pretrained ResNet-18.

\noindent\textbf{Data Splits} The fixed MultiBench directory split first assigns 4,845 samples to the non-test partition and 4,659 samples to the test set. The non-test partition is then divided using a stratified 8:2 split with seed 10, yielding 3,876 training, 969 validation, and 4,659 test samples.

\subsection*{MVSA-Single}
\noindent\textbf{Data Source and Content} MVSA-Single \cite{niu2016sentiment} contains image--text posts collected from Twitter for three-class sentiment classification: positive, neutral, and negative. The Single variant contains posts with consistent sentiment labels across annotators.

\noindent\textbf{Link} \url{https://www.kaggle.com/datasets/vincemarcs/mvsasingle}

\noindent\textbf{Feature Extraction} Images are encoded using a frozen ImageNet-pretrained ResNet-18, and text is encoded using BERT-base-uncased with attention-mask mean pooling.

\noindent\textbf{Data Splits} The provided JSONL split is used, containing 1,555 training, 518 validation, and 519 test posts.

\section{More Experimental Details}

\subsection{Supplementary Results for the Main Experiment}
Table~\ref{ap1} provides the complete quantitative results corresponding to the main experiment. For all specialized fusion methods, the reported results are averaged over three independent training runs with different random seeds ($1,2,3$), as these methods require task-specific parameter optimization. In contrast, FAM performs prediction through parameter-free multimodal in-context inference without adaptation. Therefore, its results are deterministic under the same evaluation setting and are reported directly without variance estimation.

\begin{table}[htbp]
\centering
\caption{Comparison results across different multimodal datasets. Results of the specialized fusion methods are reported as mean $\pm$ standard deviation, with the standard deviation shown below the mean in the same secondary-line font size as the $\blacktriangle$/$\blacktriangledown$ marker. The marker below each Ours ACC or AUC reports its increase/decrease relative to the mean of the corresponding metric over the specialized fusion methods.}
\label{ap1}
\setlength{\tabcolsep}{1.pt}
\resizebox{\linewidth}{!}{%
\begin{tabular}{lcccccccccccccc>{\columncolor{gray!10}}c>{\columncolor{gray!10}}c}
\toprule
\multirow{2}{*}{Dataset}
& \multicolumn{2}{c}{Concat}
& \multicolumn{2}{c}{EFT}
& \multicolumn{2}{c}{MTO}
& \multicolumn{2}{c}{OTM}
& \multicolumn{2}{c}{CAF}
& \multicolumn{2}{c}{LS}
& \multicolumn{2}{c}{TMC}
& \multicolumn{2}{>{\columncolor{gray!10}}c}{\textbf{Ours}} \\
\cmidrule(lr){2-3}
\cmidrule(lr){4-5}
\cmidrule(lr){6-7}
\cmidrule(lr){8-9}
\cmidrule(lr){10-11}
\cmidrule(lr){12-13}
\cmidrule(lr){14-15}
\cmidrule(lr){16-17}
& ACC$\uparrow$ & AUC$\uparrow$
& ACC$\uparrow$ & AUC$\uparrow$
& ACC$\uparrow$ & AUC$\uparrow$
& ACC$\uparrow$ & AUC$\uparrow$
& ACC$\uparrow$ & AUC$\uparrow$
& ACC$\uparrow$ & AUC$\uparrow$
& ACC$\uparrow$ & AUC$\uparrow$
& ACC$\uparrow$ & AUC$\uparrow$ \\
\midrule
\oursacc{GAMMA}{$\phantom{(\pm\,0.00)}$} & \oursacc{65.00}{$(\pm\,5.00)$} & \oursacc{85.59}{$(\pm\,1.58)$} & \oursacc{63.33}{$(\pm\,16.07)$} & \oursacc{74.22}{$(\pm\,15.70)$} & \oursacc{63.33}{$(\pm\,12.58)$} & \oursacc{83.11}{$(\pm\,8.20)$} & \oursacc{63.33}{$(\pm\,10.41)$} & \oursacc{76.44}{$(\pm\,15.28)$} & \oursacc{58.33}{$(\pm\,5.77)$} & \oursacc{77.48}{$(\pm\,8.23)$} & \oursacc{63.33}{$(\pm\,7.64)$} & \oursacc{77.07}{$(\pm\,17.49)$} & \oursacc{60.00}{$(\pm\,5.00)$} & \oursacc{83.56}{$(\pm\,3.11)$} & \oursacc{\textbf{85.00}}{$(\blacktriangle\,22.62)$} & \oursacc{\textbf{92.00}}{$(\blacktriangle\,12.36)$} \\
\oursacc{TCGA-KIRC}{$\phantom{(\pm\,0.00)}$} & \oursacc{71.72}{$(\pm\,4.63)$} & \oursacc{70.11}{$(\pm\,8.51)$} & \oursacc{70.71}{$(\pm\,4.63)$} & \oursacc{72.59}{$(\pm\,9.62)$} & \oursacc{71.72}{$(\pm\,4.63)$} & \oursacc{66.94}{$(\pm\,5.74)$} & \oursacc{\textbf{75.76}}{$(\pm\,9.09)$} & \oursacc{65.98}{$(\pm\,4.22)$} & \oursacc{69.70}{$(\pm\,6.06)$} & \oursacc{\textbf{73.55}}{$(\pm\,5.28)$} & \oursacc{61.62}{$(\pm\,9.74)$} & \oursacc{62.53}{$(\pm\,5.17)$} & \oursacc{63.64}{$(\pm\,3.03)$} & \oursacc{67.91}{$(\pm\,3.56)$} & \oursacc{\textbf{75.76}}{$(\blacktriangle\,6.49)$} & \oursacc{72.31}{$(\blacktriangle\,3.79)$} \\
\oursacc{NYUDv2}{$\phantom{(\pm\,0.00)}$} & \oursacc{55.45}{$(\pm\,0.64)$} & \oursacc{84.03}{$(\pm\,0.37)$} & \oursacc{30.73}{$(\pm\,2.31)$} & \oursacc{62.85}{$(\pm\,3.27)$} & \oursacc{42.86}{$(\pm\,1.79)$} & \oursacc{67.11}{$(\pm\,7.45)$} & \oursacc{52.34}{$(\pm\,3.24)$} & \oursacc{79.67}{$(\pm\,4.07)$} & \oursacc{54.18}{$(\pm\,1.39)$} & \oursacc{83.93}{$(\pm\,1.15)$} & \oursacc{55.81}{$(\pm\,1.86)$} & \oursacc{83.39}{$(\pm\,1.65)$} & \oursacc{53.57}{$(\pm\,0.64)$} & \oursacc{83.42}{$(\pm\,0.97)$} & \oursacc{\textbf{62.84}}{$(\blacktriangle\,13.56)$} & \oursacc{\textbf{90.85}}{$(\blacktriangle\,13.08)$} \\
\oursacc{CMU-MOSI}{$\phantom{(\pm\,0.00)}$} & \oursacc{57.43}{$(\pm\,1.77)$} & \oursacc{75.55}{$(\pm\,0.59)$} & \oursacc{53.60}{$(\pm\,2.62)$} & \oursacc{62.00}{$(\pm\,11.40)$} & \oursacc{67.06}{$(\pm\,1.27)$} & \oursacc{72.71}{$(\pm\,2.45)$} & \oursacc{69.83}{$(\pm\,1.77)$} & \oursacc{72.89}{$(\pm\,2.87)$} & \oursacc{67.69}{$(\pm\,1.53)$} & \oursacc{75.10}{$(\pm\,1.08)$} & \oursacc{61.61}{$(\pm\,3.08)$} & \oursacc{71.92}{$(\pm\,4.93)$} & \oursacc{65.45}{$(\pm\,1.91)$} & \oursacc{72.42}{$(\pm\,4.90)$} & \oursacc{\textbf{72.30}}{$(\blacktriangle\,9.06)$} & \oursacc{\textbf{79.58}}{$(\blacktriangle\,7.78)$} \\
\oursacc{Enrico}{$\phantom{(\pm\,0.00)}$} & \oursacc{\textbf{52.97}}{$(\pm\,0.71)$} & \oursacc{\textbf{85.65}}{$(\pm\,0.57)$} & \oursacc{26.71}{$(\pm\,2.99)$} & \oursacc{63.37}{$(\pm\,1.57)$} & \oursacc{47.15}{$(\pm\,4.85)$} & \oursacc{76.15}{$(\pm\,5.96)$} & \oursacc{45.66}{$(\pm\,1.20)$} & \oursacc{78.63}{$(\pm\,2.82)$} & \oursacc{49.77}{$(\pm\,3.33)$} & \oursacc{83.64}{$(\pm\,1.46)$} & \oursacc{50.91}{$(\pm\,4.23)$} & \oursacc{85.38}{$(\pm\,2.32)$} & \oursacc{51.94}{$(\pm\,1.62)$} & \oursacc{85.31}{$(\pm\,1.15)$} & \oursacc{52.74}{$(\blacktriangle\,6.30)$} & \oursacc{83.89}{$(\blacktriangle\,4.16)$} \\
\oursacc{ROSMAP}{$\phantom{(\pm\,0.00)}$} & \oursacc{66.67}{$(\pm\,3.93)$} & \oursacc{80.81}{$(\pm\,4.27)$} & \oursacc{63.52}{$(\pm\,10.60)$} & \oursacc{78.37}{$(\pm\,7.63)$} & \oursacc{72.96}{$(\pm\,3.57)$} & \oursacc{78.85}{$(\pm\,2.66)$} & \oursacc{70.44}{$(\pm\,4.75)$} & \oursacc{81.47}{$(\pm\,1.92)$} & \oursacc{67.92}{$(\pm\,1.63)$} & \oursacc{83.45}{$(\pm\,0.36)$} & \oursacc{61.64}{$(\pm\,9.68)$} & \oursacc{65.54}{$(\pm\,11.42)$} & \oursacc{72.96}{$(\pm\,3.31)$} & \oursacc{81.97}{$(\pm\,4.77)$} & \oursacc{\textbf{76.42}}{$(\blacktriangle\,8.40)$} & \oursacc{\textbf{84.56}}{$(\blacktriangle\,5.92)$} \\
\oursacc{UR-FUNNY}{$\phantom{(\pm\,0.00)}$} & \oursacc{60.76}{$(\pm\,0.27)$} & \oursacc{66.29}{$(\pm\,0.25)$} & \oursacc{60.83}{$(\pm\,0.52)$} & \oursacc{66.20}{$(\pm\,0.22)$} & \oursacc{60.63}{$(\pm\,0.15)$} & \oursacc{64.79}{$(\pm\,1.68)$} & \oursacc{61.67}{$(\pm\,1.32)$} & \oursacc{\textbf{66.67}}{$(\pm\,1.39)$} & \oursacc{60.73}{$(\pm\,0.25)$} & \oursacc{66.06}{$(\pm\,0.22)$} & \oursacc{59.29}{$(\pm\,1.75)$} & \oursacc{65.23}{$(\pm\,1.50)$} & \oursacc{60.66}{$(\pm\,0.46)$} & \oursacc{65.56}{$(\pm\,0.13)$} & \oursacc{\textbf{61.97}}{$(\blacktriangle\,1.32)$} & \oursacc{65.55}{$(\blacktriangledown\,0.28)$} \\
\oursacc{TCGA-SKCM}{$\phantom{(\pm\,0.00)}$} & \oursacc{54.42}{$(\pm\,1.18)$} & \oursacc{52.22}{$(\pm\,2.19)$} & \oursacc{53.74}{$(\pm\,2.36)$} & \oursacc{52.11}{$(\pm\,1.86)$} & \oursacc{56.46}{$(\pm\,5.14)$} & \oursacc{53.39}{$(\pm\,6.30)$} & \oursacc{58.50}{$(\pm\,2.36)$} & \oursacc{52.69}{$(\pm\,2.33)$} & \oursacc{52.38}{$(\pm\,10.47)$} & \oursacc{51.99}{$(\pm\,8.87)$} & \oursacc{52.38}{$(\pm\,8.50)$} & \oursacc{50.41}{$(\pm\,5.04)$} & \oursacc{56.46}{$(\pm\,7.73)$} & \oursacc{54.68}{$(\pm\,14.27)$} & \oursacc{\textbf{59.18}}{$(\blacktriangle\,4.27)$} & \oursacc{\textbf{60.53}}{$(\blacktriangle\,8.03)$} \\
\oursacc{MultiOFF}{$\phantom{(\pm\,0.00)}$} & \oursacc{58.39}{$(\pm\,1.34)$} & \oursacc{61.82}{$(\pm\,2.19)$} & \oursacc{59.96}{$(\pm\,2.71)$} & \oursacc{60.45}{$(\pm\,3.46)$} & \oursacc{61.30}{$(\pm\,3.44)$} & \oursacc{60.12}{$(\pm\,4.66)$} & \oursacc{61.07}{$(\pm\,2.93)$} & \oursacc{62.73}{$(\pm\,4.23)$} & \oursacc{61.30}{$(\pm\,0.39)$} & \oursacc{61.39}{$(\pm\,2.70)$} & \oursacc{60.40}{$(\pm\,2.42)$} & \oursacc{61.45}{$(\pm\,5.04)$} & \oursacc{61.07}{$(\pm\,4.84)$} & \oursacc{61.05}{$(\pm\,3.28)$} & \oursacc{\textbf{65.77}}{$(\blacktriangle\,5.27)$} & \oursacc{\textbf{68.44}}{$(\blacktriangle\,7.15)$} \\
\oursacc{MET-Meme(E)}{$\phantom{(\pm\,0.00)}$} & \oursacc{33.81}{$(\pm\,4.30)$} & \oursacc{\textbf{69.01}}{$(\pm\,1.34)$} & \oursacc{19.91}{$(\pm\,4.10)$} & \oursacc{54.82}{$(\pm\,3.45)$} & \oursacc{32.70}{$(\pm\,2.51)$} & \oursacc{58.78}{$(\pm\,0.12)$} & \oursacc{32.39}{$(\pm\,0.27)$} & \oursacc{62.63}{$(\pm\,1.90)$} & \oursacc{32.23}{$(\pm\,5.77)$} & \oursacc{61.07}{$(\pm\,3.92)$} & \oursacc{35.07}{$(\pm\,2.46)$} & \oursacc{66.13}{$(\pm\,4.86)$} & \oursacc{29.86}{$(\pm\,2.64)$} & \oursacc{63.39}{$(\pm\,5.70)$} & \oursacc{\textbf{37.91}}{$(\blacktriangle\,7.06)$} & \oursacc{67.02}{$(\blacktriangle\,4.76)$} \\
\oursacc{RAVDESS}{$\phantom{(\pm\,0.00)}$} & \oursacc{56.74}{$(\pm\,1.88)$} & \oursacc{90.03}{$(\pm\,0.59)$} & \oursacc{26.60}{$(\pm\,1.68)$} & \oursacc{68.77}{$(\pm\,3.31)$} & \oursacc{60.83}{$(\pm\,5.83)$} & \oursacc{87.30}{$(\pm\,3.56)$} & \oursacc{59.24}{$(\pm\,4.76)$} & \oursacc{87.79}{$(\pm\,0.87)$} & \oursacc{56.39}{$(\pm\,2.84)$} & \oursacc{88.61}{$(\pm\,0.96)$} & \oursacc{55.63}{$(\pm\,1.10)$} & \oursacc{89.90}{$(\pm\,0.52)$} & \oursacc{56.87}{$(\pm\,1.67)$} & \oursacc{89.77}{$(\pm\,0.27)$} & \oursacc{\textbf{70.00}}{$(\blacktriangle\,16.81)$} & \oursacc{\textbf{94.25}}{$(\blacktriangle\,8.23)$} \\
\oursacc{Memotion}{$\phantom{(\pm\,0.00)}$} & \oursacc{78.14}{$(\pm\,0.08)$} & \oursacc{\textbf{59.26}}{$(\pm\,0.82)$} & \oursacc{78.14}{$(\pm\,0.08)$} & \oursacc{54.69}{$(\pm\,1.88)$} & \oursacc{77.35}{$(\pm\,0.66)$} & \oursacc{56.74}{$(\pm\,2.12)$} & \oursacc{76.87}{$(\pm\,1.54)$} & \oursacc{55.42}{$(\pm\,3.42)$} & \oursacc{\textbf{78.18}}{$(\pm\,0.00)$} & \oursacc{58.13}{$(\pm\,0.46)$} & \oursacc{78.04}{$(\pm\,0.15)$} & \oursacc{57.52}{$(\pm\,1.60)$} & \oursacc{\textbf{78.18}}{$(\pm\,0.15)$} & \oursacc{59.06}{$(\pm\,0.18)$} & \oursacc{\textbf{78.18}}{$(\blacktriangle\,0.34)$} & \oursacc{57.25}{$(\blacktriangledown\,0.01)$} \\
\oursacc{Twitter1517}{$\phantom{(\pm\,0.00)}$} & \oursacc{76.15}{$(\pm\,0.93)$} & \oursacc{63.50}{$(\pm\,2.10)$} & \oursacc{75.22}{$(\pm\,0.81)$} & \oursacc{59.44}{$(\pm\,0.74)$} & \oursacc{76.22}{$(\pm\,0.16)$} & \oursacc{61.37}{$(\pm\,1.39)$} & \oursacc{76.08}{$(\pm\,0.65)$} & \oursacc{59.92}{$(\pm\,2.56)$} & \oursacc{76.01}{$(\pm\,1.01)$} & \oursacc{64.36}{$(\pm\,0.77)$} & \oursacc{76.15}{$(\pm\,0.67)$} & \oursacc{62.52}{$(\pm\,2.61)$} & \oursacc{76.43}{$(\pm\,0.38)$} & \oursacc{60.57}{$(\pm\,4.92)$} & \oursacc{\textbf{76.47}}{$(\blacktriangle\,0.43)$} & \oursacc{\textbf{65.87}}{$(\blacktriangle\,4.20)$} \\
\oursacc{Trento}{$\phantom{(\pm\,0.00)}$} & \oursacc{98.25}{$(\pm\,0.06)$} & \oursacc{99.90}{$(\pm\,0.01)$} & \oursacc{91.86}{$(\pm\,3.67)$} & \oursacc{98.01}{$(\pm\,1.29)$} & \oursacc{\textbf{98.75}}{$(\pm\,0.11)$} & \oursacc{99.93}{$(\pm\,0.06)$} & \oursacc{98.72}{$(\pm\,0.03)$} & \oursacc{\textbf{99.95}}{$(\pm\,0.02)$} & \oursacc{98.17}{$(\pm\,0.18)$} & \oursacc{99.94}{$(\pm\,0.01)$} & \oursacc{98.02}{$(\pm\,0.07)$} & \oursacc{99.87}{$(\pm\,0.02)$} & \oursacc{98.33}{$(\pm\,0.15)$} & \oursacc{99.93}{$(\pm\,0.00)$} & \oursacc{97.75}{$(\blacktriangle\,0.31)$} & \oursacc{99.83}{$(\blacktriangle\,0.18)$} \\
\oursacc{ForestNet}{$\phantom{(\pm\,0.00)}$} & \oursacc{47.36}{$(\pm\,0.91)$} & \oursacc{55.44}{$(\pm\,1.33)$} & \oursacc{46.76}{$(\pm\,0.17)$} & \oursacc{58.32}{$(\pm\,4.33)$} & \oursacc{66.52}{$(\pm\,0.57)$} & \oursacc{81.21}{$(\pm\,1.64)$} & \oursacc{60.53}{$(\pm\,4.64)$} & \oursacc{76.78}{$(\pm\,3.86)$} & \oursacc{47.31}{$(\pm\,1.50)$} & \oursacc{60.32}{$(\pm\,4.08)$} & \oursacc{47.36}{$(\pm\,1.06)$} & \oursacc{56.35}{$(\pm\,1.59)$} & \oursacc{65.37}{$(\pm\,3.72)$} & \oursacc{83.15}{$(\pm\,3.04)$} & \oursacc{\textbf{69.91}}{$(\blacktriangle\,15.45)$} & \oursacc{\textbf{86.36}}{$(\blacktriangle\,18.99)$} \\
\oursacc{MUStARD}{$\phantom{(\pm\,0.00)}$} & \oursacc{71.98}{$(\pm\,2.21)$} & \oursacc{79.52}{$(\pm\,2.93)$} & \oursacc{69.32}{$(\pm\,1.11)$} & \oursacc{75.53}{$(\pm\,2.49)$} & \oursacc{65.94}{$(\pm\,1.92)$} & \oursacc{74.48}{$(\pm\,1.66)$} & \oursacc{66.43}{$(\pm\,6.33)$} & \oursacc{72.96}{$(\pm\,2.39)$} & \oursacc{71.26}{$(\pm\,2.54)$} & \oursacc{78.08}{$(\pm\,1.49)$} & \oursacc{71.74}{$(\pm\,5.93)$} & \oursacc{78.93}{$(\pm\,2.95)$} & \oursacc{70.77}{$(\pm\,1.67)$} & \oursacc{\textbf{79.80}}{$(\pm\,0.18)$} & \oursacc{\textbf{72.46}}{$(\blacktriangle\,2.83)$} & \oursacc{78.19}{$(\blacktriangle\,1.15)$} \\
\oursacc{PAD-UFES-20}{$\phantom{(\pm\,0.00)}$} & \oursacc{74.73}{$(\pm\,0.24)$} & \oursacc{91.87}{$(\pm\,0.07)$} & \oursacc{66.02}{$(\pm\,1.70)$} & \oursacc{77.66}{$(\pm\,1.63)$} & \oursacc{71.54}{$(\pm\,0.66)$} & \oursacc{87.53}{$(\pm\,0.80)$} & \oursacc{71.86}{$(\pm\,0.16)$} & \oursacc{87.94}{$(\pm\,0.69)$} & \oursacc{72.40}{$(\pm\,0.89)$} & \oursacc{90.58}{$(\pm\,0.50)$} & \oursacc{74.38}{$(\pm\,0.61)$} & \oursacc{91.92}{$(\pm\,0.23)$} & \oursacc{74.09}{$(\pm\,0.36)$} & \oursacc{90.99}{$(\pm\,0.14)$} & \oursacc{\textbf{76.11}}{$(\blacktriangle\,3.96)$} & \oursacc{\textbf{92.51}}{$(\blacktriangle\,4.15)$} \\
\oursacc{SIIM-ISIC}{$\phantom{(\pm\,0.00)}$} & \oursacc{\textbf{97.83}}{$(\pm\,0.00)$} & \oursacc{59.67}{$(\pm\,6.23)$} & \oursacc{\textbf{97.83}}{$(\pm\,0.00)$} & \oursacc{55.88}{$(\pm\,3.97)$} & \oursacc{\textbf{97.83}}{$(\pm\,0.00)$} & \oursacc{57.55}{$(\pm\,15.52)$} & \oursacc{\textbf{97.83}}{$(\pm\,0.00)$} & \oursacc{57.34}{$(\pm\,11.89)$} & \oursacc{97.78}{$(\pm\,0.05)$} & \oursacc{66.31}{$(\pm\,21.18)$} & \oursacc{97.81}{$(\pm\,0.03)$} & \oursacc{66.55}{$(\pm\,25.69)$} & \oursacc{97.75}{$(\pm\,0.11)$} & \oursacc{57.68}{$(\pm\,13.75)$} & \oursacc{\textbf{97.83}}{$(\blacktriangle\,0.02)$} & \oursacc{\textbf{84.46}}{$(\blacktriangle\,24.32)$} \\
\oursacc{MELD}{$\phantom{(\pm\,0.00)}$} & \oursacc{62.20}{$(\pm\,0.16)$} & \oursacc{75.93}{$(\pm\,0.13)$} & \oursacc{58.10}{$(\pm\,0.83)$} & \oursacc{71.02}{$(\pm\,1.30)$} & \oursacc{64.20}{$(\pm\,1.70)$} & \oursacc{78.77}{$(\pm\,1.62)$} & \oursacc{\textbf{67.29}}{$(\pm\,0.12)$} & \oursacc{\textbf{80.77}}{$(\pm\,0.64)$} & \oursacc{65.34}{$(\pm\,0.27)$} & \oursacc{78.98}{$(\pm\,0.68)$} & \oursacc{61.76}{$(\pm\,1.03)$} & \oursacc{75.90}{$(\pm\,0.19)$} & \oursacc{63.88}{$(\pm\,0.74)$} & \oursacc{77.81}{$(\pm\,0.12)$} & \oursacc{66.21}{$(\blacktriangle\,2.96)$} & \oursacc{79.23}{$(\blacktriangle\,2.20)$} \\
\oursacc{WESAD}{$\phantom{(\pm\,0.00)}$} & \oursacc{54.68}{$(\pm\,1.13)$} & \oursacc{63.99}{$(\pm\,1.32)$} & \oursacc{71.76}{$(\pm\,3.47)$} & \oursacc{75.57}{$(\pm\,2.10)$} & \oursacc{72.64}{$(\pm\,0.37)$} & \oursacc{81.91}{$(\pm\,1.49)$} & \oursacc{71.98}{$(\pm\,1.63)$} & \oursacc{83.46}{$(\pm\,1.02)$} & \oursacc{55.66}{$(\pm\,2.46)$} & \oursacc{73.13}{$(\pm\,10.26)$} & \oursacc{55.75}{$(\pm\,2.46)$} & \oursacc{69.38}{$(\pm\,5.35)$} & \oursacc{57.43}{$(\pm\,4.30)$} & \oursacc{69.11}{$(\pm\,5.91)$} & \oursacc{\textbf{81.10}}{$(\blacktriangle\,18.26)$} & \oursacc{\textbf{91.93}}{$(\blacktriangle\,18.14)$} \\
\oursacc{Twitter2015}{$\phantom{(\pm\,0.00)}$} & \oursacc{64.77}{$(\pm\,0.79)$} & \oursacc{76.56}{$(\pm\,1.43)$} & \oursacc{63.20}{$(\pm\,1.03)$} & \oursacc{62.71}{$(\pm\,6.26)$} & \oursacc{66.47}{$(\pm\,0.92)$} & \oursacc{77.03}{$(\pm\,3.04)$} & \oursacc{67.60}{$(\pm\,0.93)$} & \oursacc{79.24}{$(\pm\,0.51)$} & \oursacc{67.12}{$(\pm\,0.93)$} & \oursacc{79.75}{$(\pm\,1.98)$} & \oursacc{64.35}{$(\pm\,0.90)$} & \oursacc{76.98}{$(\pm\,1.16)$} & \oursacc{64.51}{$(\pm\,1.67)$} & \oursacc{73.90}{$(\pm\,4.28)$} & \oursacc{\textbf{67.98}}{$(\blacktriangle\,2.55)$} & \oursacc{\textbf{81.83}}{$(\blacktriangle\,6.66)$} \\
\oursacc{SUN RGB-D}{$\phantom{(\pm\,0.00)}$} & \oursacc{\textbf{59.35}}{$(\pm\,0.55)$} & \oursacc{90.09}{$(\pm\,0.13)$} & \oursacc{47.66}{$(\pm\,4.67)$} & \oursacc{80.48}{$(\pm\,0.95)$} & \oursacc{45.83}{$(\pm\,3.90)$} & \oursacc{80.43}{$(\pm\,1.22)$} & \oursacc{58.26}{$(\pm\,2.18)$} & \oursacc{88.16}{$(\pm\,1.33)$} & \oursacc{55.49}{$(\pm\,0.53)$} & \oursacc{88.66}{$(\pm\,0.26)$} & \oursacc{58.52}{$(\pm\,0.34)$} & \oursacc{89.92}{$(\pm\,0.07)$} & \oursacc{58.02}{$(\pm\,0.25)$} & \oursacc{89.57}{$(\pm\,0.07)$} & \oursacc{57.95}{$(\blacktriangle\,3.22)$} & \oursacc{\textbf{90.26}}{$(\blacktriangle\,3.50)$} \\
\oursacc{MVSA-Single}{$\phantom{(\pm\,0.00)}$} & \oursacc{75.92}{$(\pm\,2.03)$} & \oursacc{89.83}{$(\pm\,0.57)$} & \oursacc{75.85}{$(\pm\,1.25)$} & \oursacc{83.37}{$(\pm\,2.04)$} & \oursacc{75.53}{$(\pm\,0.51)$} & \oursacc{86.70}{$(\pm\,1.24)$} & \oursacc{76.24}{$(\pm\,1.39)$} & \oursacc{87.37}{$(\pm\,1.33)$} & \oursacc{78.10}{$(\pm\,1.45)$} & \oursacc{90.41}{$(\pm\,1.39)$} & \oursacc{76.24}{$(\pm\,1.64)$} & \oursacc{90.39}{$(\pm\,0.62)$} & \oursacc{76.43}{$(\pm\,1.73)$} & \oursacc{90.48}{$(\pm\,0.15)$} & \oursacc{\textbf{78.23}}{$(\blacktriangle\,1.90)$} & \oursacc{\textbf{91.76}}{$(\blacktriangle\,3.40)$} \\
\bottomrule
\end{tabular}%
}
\end{table}

\subsection{Supplementary Results for the Effectiveness of Multimodal Extension}
Table~\ref{ap2} provides the complete comparison between TabPFN and FAM across diverse real-world datasets. Overall, our model achieves stronger performance across a broad range of multimodal prediction tasks, with improvements observed over heterogeneous modality combinations. These results demonstrate that the proposed model effectively broadens the model's capability to capture and utilize predictive correlations among heterogeneous modalities.

\begin{table}[htbp]
\centering
\caption{Comparison of TabPFN and our model across different multimodal datasets. The $\blacktriangle$/$\blacktriangledown$ marker indicates the increase/decrease of Fusion Anything relative to the corresponding TabPFN result in percentage points. Ours results that match or exceed TabPFN are shown in bold.}
\label{ap2}
\setlength{\tabcolsep}{2pt}
\begin{tabular*}{0.98\linewidth}{@{\extracolsep{\fill}}lcccccc@{}}
\toprule
\multirow{2}{*}{Dataset} & \multicolumn{3}{c}{TabPFN} & \multicolumn{3}{c}{\textbf{Ours}} \\
\cmidrule(lr){2-4}\cmidrule(lr){5-7}
& ACC$\uparrow$ & AUC$\uparrow$ & F1$\uparrow$ & ACC$\uparrow$ & AUC$\uparrow$ & F1$\uparrow$ \\
\midrule
GAMMA       & 65.00 & 80.00 & 48.48 & \oursacc{\textbf{85.00}}{$(\blacktriangle\,20.00)$} & \oursacc{\textbf{92.00}}{$(\blacktriangle\,12.00)$} & \oursacc{\textbf{78.02}}{$(\blacktriangle\,29.54)$} \\
TCGA-KIRC   & 72.73 & 79.34 & 61.18 & \oursacc{\textbf{75.76}}{$(\blacktriangle\,3.03)$} & \oursacc{72.31}{$(\blacktriangledown\,7.03)$} & \oursacc{\textbf{69.44}}{$(\blacktriangle\,8.26)$} \\
NYUDv2      & 57.80 & 87.99 & 50.51 & \oursacc{\textbf{62.84}}{$(\blacktriangle\,5.04)$} & \oursacc{\textbf{90.85}}{$(\blacktriangle\,2.86)$} & \oursacc{\textbf{57.66}}{$(\blacktriangle\,7.15)$} \\
CMU-MOSI    & 71.28 & 79.03 & 70.57 & \oursacc{\textbf{72.30}}{$(\blacktriangle\,1.02)$} & \oursacc{\textbf{79.58}}{$(\blacktriangle\,0.55)$} & \oursacc{\textbf{71.84}}{$(\blacktriangle\,1.27)$} \\
Enrico      & 51.37 & 83.16 & 37.29 & \oursacc{\textbf{52.74}}{$(\blacktriangle\,1.37)$} & \oursacc{\textbf{83.89}}{$(\blacktriangle\,0.73)$} & \oursacc{\textbf{41.25}}{$(\blacktriangle\,3.96)$} \\
ROSMAP      & 73.58 & 86.35 & 73.55 & \oursacc{\textbf{76.42}}{$(\blacktriangle\,2.84)$} & \oursacc{84.56}{$(\blacktriangledown\,1.79)$} & \oursacc{\textbf{76.36}}{$(\blacktriangle\,2.81)$} \\
UR-FUNNY    & 63.18 & 67.74 & 63.00 & \oursacc{61.97}{$(\blacktriangledown\,1.21)$} & \oursacc{65.55}{$(\blacktriangledown\,2.19)$} & \oursacc{61.78}{$(\blacktriangledown\,1.22)$} \\
TCGA-SKCM   & 55.10 & 56.32 & 54.63 & \oursacc{\textbf{59.18}}{$(\blacktriangle\,4.08)$} & \oursacc{\textbf{60.53}}{$(\blacktriangle\,4.21)$} & \oursacc{\textbf{56.09}}{$(\blacktriangle\,1.46)$} \\
MultiOFF    & 61.74 & 62.81 & 47.30 & \oursacc{\textbf{65.77}}{$(\blacktriangle\,4.03)$} & \oursacc{\textbf{68.44}}{$(\blacktriangle\,5.63)$} & \oursacc{\textbf{58.60}}{$(\blacktriangle\,11.30)$} \\
MET-Meme(E) & 33.65 & 64.40 & 18.01 & \oursacc{\textbf{37.91}}{$(\blacktriangle\,4.26)$} & \oursacc{\textbf{67.02}}{$(\blacktriangle\,2.62)$} & \oursacc{\textbf{23.87}}{$(\blacktriangle\,5.86)$} \\
RAVDESS     & 57.92 & 89.12 & 57.79 & \oursacc{\textbf{70.00}}{$(\blacktriangle\,12.08)$} & \oursacc{\textbf{94.25}}{$(\blacktriangle\,5.13)$} & \oursacc{\textbf{69.77}}{$(\blacktriangle\,11.98)$} \\
Memotion    & 78.18 & 58.20 & 43.88 & \oursacc{\textbf{78.18}}{$(\blacktriangle\,{+0.00})$} & \oursacc{57.25}{$(\blacktriangledown\,0.95)$} & \oursacc{\textbf{43.88}}{$(\blacktriangle\,{+0.00})$} \\
Twitter1517 & 76.04 & 62.24 & 44.06 & \oursacc{\textbf{76.47}}{$(\blacktriangle\,0.43)$} & \oursacc{\textbf{65.87}}{$(\blacktriangle\,3.63)$} & \oursacc{\textbf{44.21}}{$(\blacktriangle\,0.15)$} \\
Trento      & 98.58 & 99.91 & 97.85 & \oursacc{97.75}{$(\blacktriangledown\,0.83)$} & \oursacc{99.83}{$(\blacktriangledown\,0.08)$} & \oursacc{96.11}{$(\blacktriangledown\,1.74)$} \\
ForestNet   & 68.86 & 88.63 & 64.48 & \oursacc{\textbf{69.91}}{$(\blacktriangle\,1.05)$} & \oursacc{86.36}{$(\blacktriangledown\,2.27)$} & \oursacc{\textbf{65.94}}{$(\blacktriangle\,1.46)$} \\
MUStARD     & 73.19 & 79.81 & 73.19 & \oursacc{72.46}{$(\blacktriangledown\,0.73)$} & \oursacc{78.19}{$(\blacktriangledown\,1.62)$} & \oursacc{72.46}{$(\blacktriangledown\,0.73)$} \\
PAD-UFES-20 & 76.50 & 93.15 & 62.71 & \oursacc{76.11}{$(\blacktriangledown\,0.39)$} & \oursacc{92.51}{$(\blacktriangledown\,0.64)$} & \oursacc{\textbf{64.04}}{$(\blacktriangle\,1.33)$} \\
SIIM-ISIC   & 97.83 & 83.69 & 49.45 & \oursacc{\textbf{97.83}}{$(\blacktriangle\,{+0.00})$} & \oursacc{\textbf{84.46}}{$(\blacktriangle\,0.77)$} & \oursacc{\textbf{49.45}}{$(\blacktriangle\,{+0.00})$} \\
MELD        & 66.82 & 78.35 & 61.44 & \oursacc{66.21}{$(\blacktriangledown\,0.61)$} & \oursacc{\textbf{79.23}}{$(\blacktriangle\,0.88)$} & \oursacc{60.81}{$(\blacktriangledown\,0.63)$} \\
WESAD       & 81.62 & 93.23 & 74.82 & \oursacc{81.10}{$(\blacktriangledown\,0.52)$} & \oursacc{91.93}{$(\blacktriangledown\,1.30)$} & \oursacc{72.66}{$(\blacktriangledown\,2.16)$} \\
Twitter2015 & 66.73 & 80.95 & 56.14 & \oursacc{\textbf{67.98}}{$(\blacktriangle\,1.25)$} & \oursacc{\textbf{81.83}}{$(\blacktriangle\,0.88)$} & \oursacc{55.97}{$(\blacktriangledown\,0.17)$} \\
SUN RGB-D   & 58.70 & 90.93 & 55.44 & \oursacc{57.95}{$(\blacktriangledown\,0.75)$} & \oursacc{90.26}{$(\blacktriangledown\,0.67)$} & \oursacc{\textbf{57.41}}{$(\blacktriangle\,1.97)$} \\
MVSA-Single & 79.58 & 91.60 & 76.76 & \oursacc{78.23}{$(\blacktriangledown\,1.35)$} & \oursacc{\textbf{91.76}}{$(\blacktriangle\,0.16)$} & \oursacc{74.76}{$(\blacktriangledown\,2.00)$} \\
\bottomrule
\end{tabular*}
\end{table}

\end{document}